\RequirePackage{fix-cm}
\documentclass[smallextended,natbib]{svjour3}   
\smartqed  
\usepackage{graphicx}
\usepackage{multirow}
\usepackage[table,xcdraw]{xcolor}
\usepackage{multicol}
\usepackage{comment}
\usepackage{marvosym}
\usepackage{multibib}
\usepackage{microtype}
\usepackage{times}
\usepackage{xspace}
\usepackage{tabularx}
\usepackage{morefloats}
\usepackage{epstopdf}
\usepackage[T1]{fontenc}
\usepackage[utf8x]{inputenc}
\usepackage{textcomp}
\usepackage{booktabs}
\usepackage{pdflscape}
\usepackage{lipsum}
\usepackage{footnote}

\usepackage{geometry}
\usepackage{appendix}
\usepackage{hyperref}
\usepackage{xstring}
\usepackage{tikz-dependency}

\usepackage{wrapfig}

\usepackage{examples} 

\newcommand\A{ATDT\xspace}
\newcommand\B{Hi-En-CS\xspace}
\newcommand\C{TAAE\xspace}
\newcommand\D{TWRO\xspace}
\newcommand\E{DWT\xspace}
\newcommand\rF{W2.0\xspace}
\newcommand\rG{Frb\xspace}
\newcommand\rH{Twb\xspace}
\newcommand\rI{Twb2\xspace}
\newcommand\rJ{TDT\xspace}
\newcommand\rK{xUGC\xspace}
\newcommand\rL{ITU\xspace}
\newcommand\rM{tweeDe\xspace}
\newcommand\rN{Pst\xspace}
\newcommand\rO{FSMB\xspace}
\newcommand\rP{EWT\xspace}
\newcommand\rQ{STB\xspace}
\newcommand\rR{CWT\xspace}
\newcommand\rS{GUM\xspace}
\newcommand\rT{MNo\xspace}
\newcommand\NBZ{NBZ\xspace}

\definecolor{babyblueeyes}{rgb}{0.63, 0.79, 0.95}
\definecolor{bittersweet}{rgb}{1.0, 0.44, 0.37}
\definecolor{bleudefrance}{rgb}{0.19, 0.55, 0.91}
\definecolor{applegreen}{rgb}{0.55, 0.71, 0.0}
\definecolor{antiquefuchsia}{rgb}{0.57, 0.36, 0.51}
\definecolor{asparagus}{rgb}{0.53, 0.66, 0.42}

\newcommand\citelanguageresource\cite

\newcommand\newcite\citet

\expandafter\def\expandafter\UrlBreaks\expandafter{\UrlBreaks
  \do\a\do\b\do\c\do\d\do\e\do\f\do\g\do\h\do\i\do\j%
  \do\k\do\l\do\m\do\n\do\o\do\p\do\q\do\r\do\s\do\t%
  \do\u\do\v\do\w\do\x\do\y\do\z\do\A\do\B\do\C\do\D%
  \do\E\do\F\do\G\do\H\do\I\do\J\do\K\do\L\do\M\do\N%
  \do\O\do\P\do\Q\do\R\do\S\do\T\do\U\do\V\do\W\do\X%
  \do\Y\do\Z}

\usepackage{soulutf8}
\usepackage[normalem]{ulem}

\usepackage{enumitem}
\setlist{nolistsep}
\usepackage{microtype}

\iftrue 
    \usepackage[final]{proofing}
  \renewcommand\hl[1]{{#1}}  
   \newcommand\draftHL[2]{#1}{\draftnote{\red{#2}}}
   \newcommand\redHL[1]{}
  \newcommand\todo[1]{}

  \newcommand{\Djame}[1]{}

\else  

\usepackage[draft]{proofing}

\newcommand{\Djame}[1]{
\textbf{\textcolor{red}{\hl{Djame: #1}}}
}

\newcommand\red[1]{{\textbf{\textcolor{red}{#1}}}}

\newcommand\draftHL[2]{\hl{#1}{\draftnote{\red{#2}}}}

\let\oldred\red
\renewcommand\red[1]{{\bf \oldred{{#1}}}}

 \renewcommand\draftHL[2]{\hl{#1}{\draftnote{\red{#2}}}}
 \newcommand\redHL[1]{\red{\hl{#1}}}
\let\olddraftnote\draftnote
\renewcommand\draftnote[1]{\olddraftnote{\red{#1}}}

\fi

\newcommand{\treetrans}[1]{%
  \node[anchor=north,yshift=-1ex,font=\small] at (\matrixref.south) {#1};%
}

\newcommand{\treetransbelow}[1]{%
  \node[anchor=north,yshift=-5ex,font=\small] at (\matrixref.south) {#1};%
}
\newcommand{\twsource}[1]{(from Twitter, #1)}
\newcommand{\twadapt}[1]{(adapted from Twitter, #1)}

\begin{document}

\title{Treebanking User-Generated Content
}

\subtitle{A UD Based Overview of Guidelines, Corpora and Unified Recommendations}


\author{Anonymous submission}
 \author{Manuela Sanguinetti${^1}$
         \and Lauren Cassidy$^{3}$
         \and Cristina Bosco$^{2}$
         \and Özlem Çetinoğlu$^{4}$
         \and Alessandra Teresa Cignarella$^{2,5}$
         \and Teresa Lynn$^{3}$
         \and Ines Rehbein$^{6}$
         \and Josef Ruppenhofer$^{7}$
         \and Djamé Seddah$^{8}$
         \and Amir Zeldes$^{9}$
 }

\authorrunning{Sanguinetti et al.} 

 \institute{1. Dipartimento di Matematica e Informatica, Università degli Studi di Cagliari, Italy \and 
           2. Dipartimento di Informatica, Università degli Studi di Torino, Italy \and 
           3. ADAPT Centre, Dublin City University, Ireland \and 
           4. IMS, University of Stuttgart, Germany \and 
           5. PRHLT Research Center, Universitat Politècnica de València, Spain \and 
           6. University of Mannheim, Germany \and 
           7. Leibniz-Institut für Deutsche Sprache Mannheim, Germany \and 
           8. INRIA Paris, France \and 
           9. Georgetown University, USA. 
           \textit{Corresponding author}: Manuela Sanguinetti (\texttt{manuela.sanguinetti@unica.it}).}

\date{Received: date / Accepted: date}

\maketitle
\begin{abstract}
This article presents a discussion on the main linguistic phenomena which cause difficulties in the analysis of user-generated texts found on the web and in social media, and proposes a set of annotation guidelines for their treatment within the Universal Dependencies (UD) framework of syntactic analysis. Given on the one hand the increasing number of treebanks featuring user-generated content, and its somewhat inconsistent treatment in these resources on the other, the aim of this article is twofold: (1) to provide a condensed, though comprehensive, overview of such treebanks -- based on available literature -- along with their main features and a comparative analysis of their annotation criteria, and (2) to propose a set of tentative UD-based annotation guidelines, to promote consistent treatment of the particular phenomena found in these types of texts. The overarching goal of this article is to provide a common framework for researchers interested in developing similar resources in UD, thus promoting cross-linguistic consistency, which is a principle that has always been central to the spirit of UD. 

\end{abstract}


\section{Introduction}

The immense popularity gained by social media in the last decade has made it an attractive source of data for a large number of research fields and applications, especially for sentiment analysis and opinion mining \citep{Balahur2013,SeverynaEtAl2016}. In order to successfully process the data available from such sources, linguistic analysis is often helpful \citep{VilaresEtAl2017,MataouiEtAl2018}, which in turn prompts the use of NLP tools to that end. Despite the ever increasing number of contributions, especially on Part-of-Speech tagging \citep{gimpel-etal-2011-part,Owoputi2013,lynn-2015,Bosco2016,cetinoglu:2016b,Proisl2018,rehbein:etal:2018,BehzadZeldes2020} and parsing \citep{foster:2010:cba,Petrov2012,kong:etal:2014,Liu2018,Sanguinetti2018a}, automatic processing of user-generated content (UGC) still represents a challenging task, as is shown by some tracks of the workshop series on noisy user-generated text (W-NUT).\footnote{\url{https://noisy-text.github.io/}.}
UGC is a continuum of text sub-domains that vary considerably according to the specific conventions and limitations posed by the medium used (blog, discussion forum, online chat, microblog, etc.), the degree of ``canonicalness'' with respect to a more standard language, as well as the linguistic devices\footnote{This phrase is used here in a broad sense to indicate all those orthographic, lexical as well as structural choices adopted by a user, often for expressive purposes.} adopted to convey a message. 
Overall, however, there are some well-recognized phenomena that characterize UGC as a whole \citep{foster:2010:cba,Seddah2012p,eisenstein-2013-bad}, and that continue to make its treatment a difficult task.

As the availability of \textit{ad hoc} training resources remains an essential factor for the analysis of these texts, the last decade has seen numerous resources of this type being developed. A good proportion of those resources that contain syntactic analyses have been annotated according to the Universal Dependencies (UD) scheme \citep{UD2016}, a dependency-based annotation scheme which has become a popular standard reference for treebank annotation because of its adaptability to different domains and genres. At the time of writing (in UD version 2.6), as many as 92 languages are represented within this vast project, with 163 treebanks dealing with extremely varied genres, ranging from news to fiction, medical, legal, religious texts, etc. This linguistic and textual variety demonstrates the generality and adaptability of the annotation scheme.

On the one hand, this flexibility opens up the possibility of also adopting the UD scheme for a broad range of user-generated text types, since a framework which is proven to be readily adaptable is more likely to fit the needs of diverse UGC data sources, and the wealth of existing materials makes it potentially easier to find precedents for analysis whenever difficult or uncommon constructions are encountered. On the other hand, the current UD guidelines do not fully account for some of the specifics of UGC domains, thus leaving it to the discretion of the individual annotator (or teams of annotators) to interpret the guidelines and identify the most appropriate representation of these phenomena. This article therefore draws attention to the annotation issues of UGC, 
while attempting to find a cross-linguistically consistent representation, all within a single coherent framework. It is also worth pointing out that inconsistencies may be found even among multiple resources in the same language (see e.g. \citep{aufrant2017limsi,bjorkelund-etal-2017-ims})\footnote{The different French resources currently available in the repository are also an example of such inconsistencies, which are mostly due to different annotation choices inherited from different linguistic traditions (e.g. Fr\_ParTUT vs Fr\_Sequoia) or different annotator teams (e.g. SynTagRus vs GSD\_Russian).}.
Therefore, even on the level of standardizing a common solution for UGC and other treebanks in one language, some more common guidance taking UGC phenomena into account is likely to be useful. 
This article first 
provides an overview of the existing resources -- treebanks in particular -- of user-generated texts from the Web, with a focus on comparing their varying annotation choices with respect to certain phenomena typical of this domain. Next, we  present a systematic analysis of some of these phenomena within the context of the framework of UD, surveying previous solutions, and propose, where possible, guidelines aimed at overcoming the inconsistencies found among the existing resources.

Given the nature of the phenomena covered and the fact that the existing relevant resources only cover a handful of languages, we are aware that the debate on their annotation is still wide open; this article therefore has no prescriptive intent. That said, the proposals in this article represent the consensus of a fairly large group of UD contributors working on diverse languages and media, with the goal of building a critical mass of resources that are annotated in a consistent way. As such, it can be used as a reference when considering alternative solutions, and it is hoped that the survey of treatments of similar phenomena across resources will help future projects in making choices that are as comparable as possible to common practices in the existing datasets.

\section{Linguistics of UGC}
\label{sec:lingUGC}
Describing all challenges brought about by UGC for all languages is beyond the scope of this work. Nevertheless, following
\newcite{foster:2010:cba,Seddah2012p,eisenstein-2013-bad} 
we can characterize UGC's idiosyncrasies along a few major dimensions defined by the intentionality or communicative needs that motivate linguistic variation. It should be stressed that one and the same utterance, and indeed often a single word, can instantiate multiple categories from the selection below, and that their occurrence can be either intentional or unintentional.\footnote{In fact, because of the inherent uncertainty in interpreting corpus utterances, coupled with the often highly contextual nature of UGC, it is important to apply analyses that are as independent as possible from definitions referring to speaker or writer intentions.}

\begin{figure}[!ht]
\centering
\includegraphics[width=0.85\textwidth]{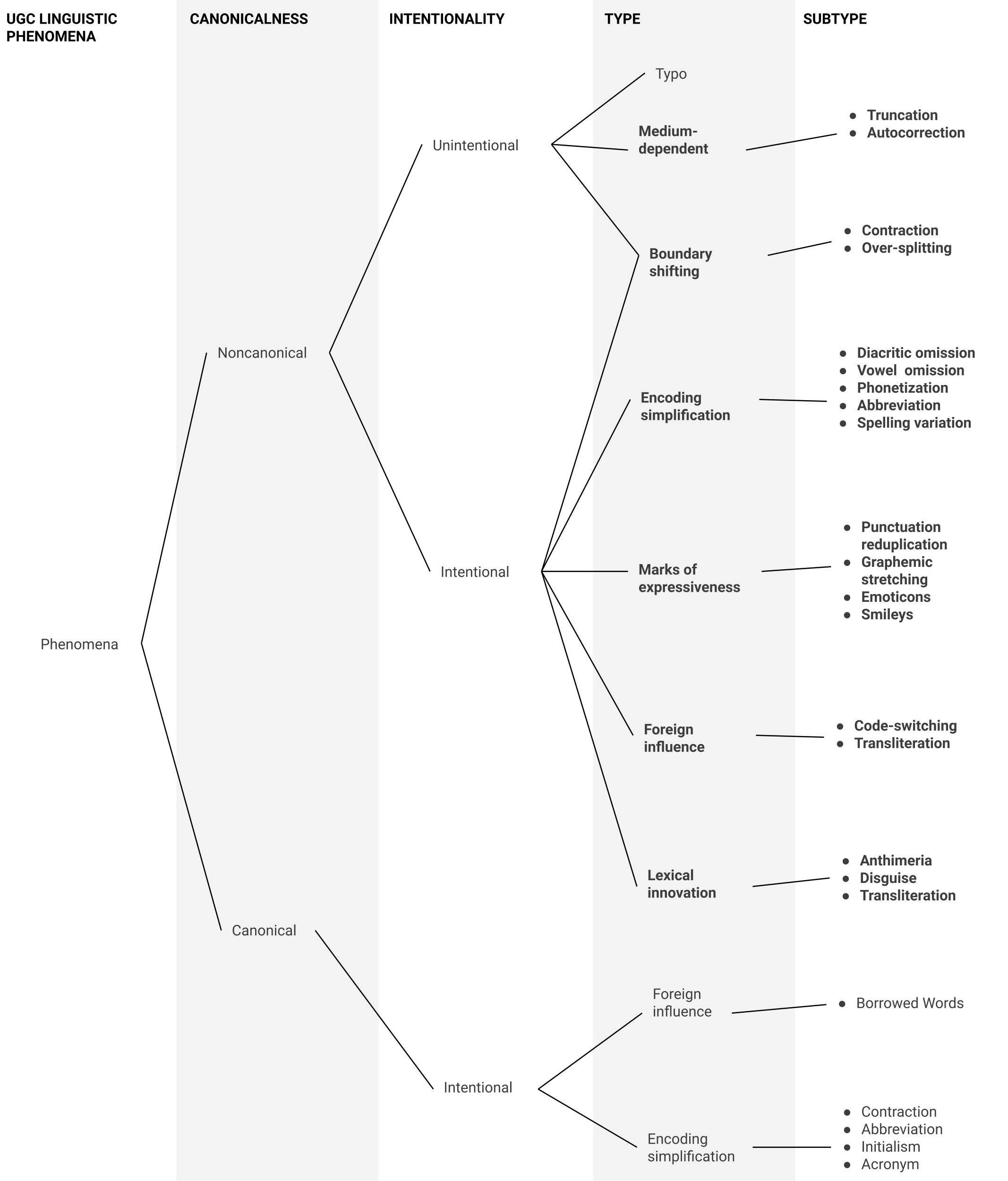}
\caption{Diagram of UGC linguistic phenomena. Our focus is on the noncanonical linguistic phenomena prevalent in UGC which do not yet have standardized annotation guidelines within the UD framework. These elements in boldface are exemplified in Table \ref{tab:UGCex}. }\label{fig:diagram_ugc_phenomena}
\end{figure}

Figure \ref{fig:diagram_ugc_phenomena} displays the hierarchy we followed to describe UGC phenomena. “Canonicalness” refers to whether a phenomenon is also observable in standard text. “Intentionality” refers to whether its production was deliberate.\footnote{This categorization may only be guessed at by the annotator as it is unknowable by observing the surface text alone.} “Type” refers to the variety of the phenomenon, while "Subtype" provides sub-categorization of each type. 

\begin{itemize} 
\item \textbf{Encoding simplification:} This axis covers ergographic phenomena, i.e. phenomena aiming to reduce the effort of writing, such as diacritic or vowel omissions (EN {\em people} $\rightarrow$ {\em ppl}).
 \item {\bf Boundary shifting}: 
 Some phenomena affect the number of tokens, compared to standard orthography, either by replacing several standard language tokens by only one, which we will refer to as \texttt{contraction} 
 (FR {\em n'importe} $\rightarrow$ {\em nimp} `whatever, rubbish') or conversely by splitting one standard language token into several tokens, which we will refer to as \texttt{over-splitting} (FR {\em c'était} $\rightarrow$ {\em c t}, `it was'). In some cases, the resulting non-standard tokens might even be homographs of existing words, creating more ambiguities if not properly analyzed. Such phenomena are frequent in the corpora of UGC surveyed below, and they require specific
  annotation guidelines. 
  
\item \textbf{Marks of expressiveness}: orthographic variation is often used as a mark of
  expressiveness, e.g., graphical stretching ({\em yes} $\rightarrow$ {\em yesssss}),
  replication of punctuation marks ({\em ?} $\rightarrow$ {\em ?????}), as well as emoticons, which can also take the place of standard language words, e.g. a noun or verb (FR {\em Je t'aime} $\rightarrow$ {\em Je t'<3}, `I love you', with the heart emoticon representing the verb `love'). These phenomena often emulate sentiment expressed through prosody, facial expression and gesture in direct interaction; however the written nature of UGC data means that they need to be assigned analyses in terms of tokens, parts of speech and dependency functions. Many of the symbols involved also contain punctuation, which can lead to spurious tokenization and problems in lemmatization (see below).

\item {\bf Foreign language influence}: 
UGC is often produced in highly multilingual settings and we often find evidence of the influence of foreign language(s) on the users' text productions, especially in code-switching scenarios, in domain-specific conversations (video game chat logs) or in the productions of L2 speakers, all of which complicate the typically monolingual context for which syntactic annotation guidelines are developed. In some cases, foreign words are imported as is from a donor language (e.g. IT {\em non fare la bad girl} `don't be a bad girl' instead of {\em non fare la cattiva ragazza}). In other cases, foreign influence can create novel words: a good example is an Irish term coined by one user to mean `awkward', {\em áicbheaird}, whose pronunciation mimics the English
word (instead of the equivalent standard Irish term {\em amscaí}).

\item {\bf Medium-dependent phenomena:} Some deviations from standard language are direct results of the electronic medium, including client-side automatic error correction, masking or replacement of taboo words by the server, artifacts of the keyboard or other user input devices, and more. In some cases, and especially for languages other than English, some apparent English words in UGC represent automatic `corrections' of non-English inputs, such as Irish {\em coic\'{i}se} `fortnight' $\rightarrow$ {\em concise}. These cases raise questions relating to the degree of interpretation, such as reconstructing likely UGC inputs before error correction, which may need to be annotated either as typos (in UD, the annotation \texttt{Typo=Yes}), or at an even greater level of detail in lemmatization.

\item {\bf Context dependency:} Given the conversational nature of most social media, UGC data often exhibits high context-dependence (much like  dialogue-based interaction). 
Speaker turns in UGC are often marked by the thread structure in a Web interface or app, and information from across a thread may provide a rich context for varying levels of ellipsis and anaphora that are much less frequent or complex in standard written language. In addition, multimedia content, pictures or game events can serve as a basis for discussion and are used as external context points, acting, so to speak, as non-linguistic antecedents or targets for deixis and argument structure. This can make the annotation task more difficult and prone to interpretation errors, especially if the actual thread context is not available, and mandates some conventional guidelines.
\end{itemize}


 

\noindent Table \ref{tab:UGCex} presents some cross-language examples of several of the phenomena outlined above. The elements in boldface in Figure~\ref{fig:diagram_ugc_phenomena} are exemplified in Table \ref{tab:UGCex}.

\begin{table*}[!ht]
\resizebox{\textwidth}{!}{%
\centering
\begin{tabular}{lllll}
\toprule
{\bf Phenomenon} & {\bf Lang} & {\bf Attested example} & {\bf Standard form} &
{\bf Gloss}\\\midrule
\multicolumn{5}{c}{\sc \textcolor{bleudefrance}{\bf Encoding Simplification}}\\\midrule
\textcolor{bleudefrance}{\bf Diacritic Omission}  & GA & {\em {\bf Leigh aris}!} & {\em Léigh arís!} & `Read again!'\\
  & TR & {\em {\bf Istanbuldaki agaclar}} & {\em İstanbul'daki ağaçlar} & `trees in Istanbul'\\
\textcolor{bleudefrance}{\bf Vowel Omission} & EN & {\em {\bf ppl}} & {\em people} & `people'\\
 & TR & {\em {\bf slm}} & {\em selam} & `hi'\\
\textcolor{bleudefrance}{\bf Phonetization}& EN & {\em Happy Birthday {\bf 2} me} & {\em Happy Birthday to me} & `Happy Birthday to me'\\
  & TR & {\em {\bf 1}az} & {\em biraz} & `some'\\
  & DE & {\em  k{\bf 1} Mensch hat so} & {\em kein Mensch hat so} & `nobody has such a\\ 
    &   & {\em  {\bf 1} Thailandhass} & {\em einen Thailandhass} & hatred of Thailand'\\ 

\textcolor{bleudefrance}{\bf Spelling Variation} & FR & {\em je {\bf sé}} & {\em je sais} & `I know'\\
& GA & {\em {\bf gura} míle} & {\em go raibh míle} & `thank you very much'\\

& FR & {\em tous mes {\it\bf examen}} & {\em
 tous mes examens} & `All my examinations\\
&  & {\em {\bf son} normaux} & {\em sont normaux} & are normal'\\
& IT & {\em {\it\bf anno} mangiato} & {\em hanno mangiato} & `(they) have eaten'\\
\textcolor{bleudefrance}{\bf Abbreviation}  & EN & {\em {\bf govt}} & {\em government} & `government'\\
& DE& {\em {\bf zuggm}} & {\em zugegebenermaßen} & `admittedly'\\
\midrule
\multicolumn{5}{c}{\sc \textcolor{bittersweet}{\bf Boundary Shifting}}\\\midrule
\textcolor{bittersweet}{\bf Contraction}
& FR & {\em {\bf nimp} quoi} & {\em n'importe quoi} & `rubbish'\\

\textcolor{bittersweet}{\bf Over-splitting}
& FR & {\em {\bf c a dire}} & {\em c'est-à-dire} & `namely'\\
\textcolor{red}{\bf ~~}
& TR & {\em {\bf gele bilirim}} & {\em gelebilirim} & `I can come'\\
\midrule
\multicolumn{5}{c}{\sc \textcolor{asparagus}{\bf Marks of Expressiveness}}\\\midrule
\textcolor{asparagus}{\bf Punct. reduplication}
& FR &{\em Joli~\textbf{!!!!!!}} & {\em Joli~!} & `nice!'\\
& IT &{\em chi\textbf{?!?!?!}} & {\em chi?} & `who?'\\
\textcolor{asparagus}{\bf Graphemic stretching}
& EN & {\em supe\textbf{rrrrrrrrr}} & {\em super} & `great'\\
& IT & {\em \textbf{siiiiiiiiiiiii}} & {\em sì} & `yes'\\

\textcolor{asparagus}{\bf  Emoticons/smileys}
& -  & {\em :-) {\em <3}} & -- & --\\
& GA & {\em \textbf{<3} mór} & {\em Grá mór} & `Lots of love'\\
\midrule

\multicolumn{5}{c}{\sc \textcolor{antiquefuchsia}{\bf {Lexical Innovation}}}\\\midrule
\textcolor{antiquefuchsia}{\bf Disguise} & IT & {\em {\it\bf caxxo}} & {\em cazzo} & `fuck'\\
  & TR & {\em {\it\bf mok / b.k / b*k}} & {\em bok} & `shit'\\
  & DE & {\em \textbf{Verfi**t} lange Reise} & {\em Verfickt lange Reise} & `fucking long trip'\\
\textcolor{antiquefuchsia}{\bf Transliteration}
& GA & {\em {\bfáicbheaird}} & {\em amscaí} & `awkward'\\
& TR & {\em{\bf taymlayn}} & {\em zaman akışı} & `timeline'\\

\textcolor{antiquefuchsia}{\bf Anthimeria} 
& IT & {\em{\bf tuittare}} & {\em twittare} & `to tweet'\\
& EN & feel free to {\em PM} & {\em personal message} & `to send a message'\\
& DE & {\em{\bf achtisch}} & {\em EN eightish} & `about 8 o'clock'\\
\midrule
\multicolumn{5}{c}{\sc \textcolor{orange}{\bf Medium-dependent Phenomena}}\\\midrule
\textcolor{orange}{\bf Truncation}
& GA & {\em thart fa' 53 {\bf nó…}} & {\em thart fa' 53 nóiméad} & `over 53 mi…(minutes)'\\
\textcolor{orange}{\bf Autocorrection} &
 GA & {\em{\bf concise}} & {\em coicíse} & `fortnight'\\
\bottomrule
\end{tabular}
}
\caption{Multi-lingual examples of UGC phenomena (DE:German, EN:English, FR:French, GA:Irish, IT:Italian, TR:Turkish).}\label{tab:UGCex}
\end{table*}

\section{UGC Treebanks: An Overview}
\label{sec:web}
In order to provide an account of the resources described in the literature, we carried out a semi-systematic search on Google Scholar using the following set of keywords (\textit{treebank web social media}) and (\textit{universal dependencies web social media}), limiting to the first five pages, sorted by relevance, and without time filters. 
We selected only open-access papers describing either a novel resource or an already-existing one that has been expanded or altered in such a way that it gained the status of a new one. 
In the few cases of multiple publications referring to the same resource, we chose the most recent one, assuming it contained the most up-to-date information regarding the status of the resource.
We also included in our collection five papers that we were aware of, but which were not retrieved by the search. As the main focus of this work is on the syntactic annotation of web content and user-generated texts, we discarded all papers that presented system descriptions, parsing experiments or POS-tagged resources (without syntactic annotation).
The results of our search are summarized in Table \ref{tab:overview}.

\begin{savenotes}
\begin{table*}[t]
\begin{footnotesize}
\centering
\begin{tabular}{llllc}
 \textbf{Name} & \textbf{Reference} & \textbf{Source} & \textbf{Language}& \textbf{UD-based} \\
\toprule
 ATDT (UD) & \citep{Albogamy2017a} & Twitter & AR & yes\\ \midrule
Hi-En-CS & \citep{Bhat2018a} & Twitter & HI/EN & yes \\ \midrule
 \multirow{2}{*}{TwitterAAE (TAAE)} & \multirow{2}{*}{\citep{Blodgett2018}} & \multirow{2}{*}{Twitter} & AAE, & \multirow{2}{*}{yes}\\ 
  & & & MAE & \\ \midrule
 TWITTIRÒ-UD (TWRO) & \citep{Cignarella2019} & Twitter & IT & yes\\ \midrule
 DWT & \citep{Daiber2016} & Twitter & EN & no\textsuperscript{$\star$}\\ \midrule
 \multirow{2}{*}{W2.0} & \multirow{2}{*}{\citep{Foster2011}} & Twitter,  & \multirow{2}{*}{EN} & \multirow{2}{*}{no\textsuperscript{$\ddagger$}}\\ 
   &  & sport fora &  & \\ \midrule
 Foreebank (Frb) & \citep{Kaljahi2015} & technical fora & EN, FR & no\textsuperscript{$\ddagger$}\\ \midrule
 Tweebank (Twb) & \citep{kong:etal:2014} & Twitter & EN & no\textsuperscript{$\star$}\\ \midrule
 Tweebank2 (Twb2) & \citep{Liu2018} & Twitter & EN & yes\\ \midrule
 TDT & \citep{Luotolahti2015} & various & FI & yes\\ \midrule
 xUGC & \citep{Martinez2016} & various & FR & yes\\ \midrule
 ITU & \citep{Pamay2015} & n.a. & TR & no\textsuperscript{$\star$}\\ \midrule
 tweeDe & \citep{Rehbein2019} & Twitter & DE & yes\\ \midrule
 PoSTWITA-UD (Pst) & \citep{Sanguinetti2018a} & Twitter & IT & yes\\ \midrule
 FSMB & \citep{Seddah2012p} & Twitter, Facebook& FR & no\textsuperscript{$\ddagger$}\\ 
  &  & discussions fora&& \\ 
\midrule
 Narabizi (NBZ) & \citep{seddah-etal-2020-building} & Newspaper fora & DZ/FR & yes \\
 \midrule
 EWT & \citep{Silveira2014} & various & EN & yes\\ \midrule
 MoNoise (MNo) & \citep{van2018modeling} & Twitter & EN & yes \\  \midrule
 STB & \citep{Wang2017} & discussion fora & SgE & yes\\ \midrule
 \multirow{2}{*}{CWT} & \multirow{2}{*}{\citep{Wang2014}} & Twitter, & \multirow{2}{*}{ZH} & \multirow{2}{*}{no\textsuperscript{$\star$}}\\
   & & Sina Weibo &  & \\\midrule
 GUM & \citep{Zeldes2017gum} & various & EN & yes \\
\bottomrule
\end{tabular}
\caption{Overview of treebanks featuring user-generated content from the web, along with some basic information on the data source, the languages involved and whether they are based on UD scheme or not. In non-UD treebanks, $\ddagger$ and $\star$ indicate, respectively, a constituency or dependency-based syntactic representation.
(AR:Arabic,
DZ/FR: Dialectal North-African Arabic/French code-switching,
HI/EN:Hindi-English code-switching,
AAE:African-American English,
MAE:Mainstream American English,
IT:Italian,
EN:English,
FR:French,
FI:Finnish,
TR:Turkish,
DE:German,
SgE:Singapore English,
ZH:Chinese).}\label{tab:overview}
\end{footnotesize}
\end{table*}
\end{savenotes}

Based on the selection criteria mentioned above, we found 21 papers describing a resource featuring web/social media texts; most of them are freely available, either from a GitHub/BitBucket repository, a dedicated web page or upon request. Dataset sizes vary widely, ranging from 500 (\E) to approximately 6,700 tweets (\rN) for the Twitter treebanks, and from 974 (\rK) to more than 16,000 sentences (\rP) for the other datasets. 

\subsection{Languages} English is the most represented language, however, some of the resources focus on different English language varieties such as African-American English (\C), Singaporean English (\rQ), and Hindi-Indian English code-switching data (\B). Three resources are in French (\rG, \rK, \rO), one includes code-switching data in French and transliterated dialectal North-African Arabic (\NBZ) and two in Italian (\D, \rN); the remaining ones are in Arabic (\A), Chinese (\rR), Finnish (\rJ), German (\rM) and Turkish (\rL). While the current Irish Twitter corpus, which is the source of some of the examples in Table \ref{tab:UGCex}, has not yet been converted to treebank format (and as such is not listed in Table~\ref{tab:overview}), its annotation  presented most of the same challenges that make up the discussion below \citep{lynn-2015,lynn-2019}.

\subsection{Data sources} 13 out of 21 resources are either partially or entirely made up of Twitter data. Possible reasons for this are the easy retrieval of the data by means of the Twitter API and by the use of wrappers for crawling the data, as well as the policy adopted by the platform with regard to the use of data for academic and non-commercial purposes.\footnote{\url{https://developer.twitter.com/en/developer-terms/agreement-and-policy\#c-respect-users-control-and-privacy}} Only three resources include data from social media other than Twitter, specifically Facebook (\rO), Reddit (\rS) and Sina Weibo (\rR), and, overall, most of the remaining resources comprise texts from discussion fora of various kinds. Only three treebanks consist of texts from different sub-domains, i.e. newspaper fora (\NBZ), blogs, reviews, emails, newsgroups and question answers (\rP), and Wikinews, Wikivoyage, wikiHow, Wikipedia biographies, interviews, academic writing, Creative Commons fiction (\rS). One resource is made up of generic data automatically crawled from the web (\rJ).

\subsection{Syntactic frameworks}

With regard to the formalism adopted to represent the syntactic structure, dependencies are by far the most used paradigm, especially among the treebanks created from 2014 onwards, though some resources include both constituent and dependency syntax versions -- \rP has manually annotated constituent trees, while \rS contains automatic constituent parses based on parser output from CoreNLP \citep{ManningEtAl2014} applied to the gold POS tags. As pointed out by \newcite{Martinez2016},  dependency-based annotation lends itself well to noisy texts, since it is easier to deal with disfluencies and fragmented text breaking conventional phrase structure rules, which prohibit discontinuous constituents.

The increasing popularity of UD may also have a role in the prevalence of dependencies for web data, considering that 14 out of the 18 dependency treebanks are based on the UD scheme. Although not all of these corpora have been released in the official UD repository, and some of them do not strictly comply with the latest format specifications, the large number of UD 
resources, as well as their occasional divergences, highlight the need 
to converge on a single syntactic annotation framework for UGC within UD, to allow for a better degree of comparability across the resources and arrive at tested best practices. 

In the next section, we provide an analysis of the guidelines of the surveyed treebanks, highlighting their similarities and differences, and 
a preliminary classification of the phenomena to be dealt with in UGC data from social media and the web with respect to the standard grammar framework for each language.

\subsection{Annotation Comparison}\label{ssec:comparison}
To explore the similarities and divergences among the resources summarized in Table \ref{tab:overview}, we carried out a comparative analysis of recurring annotation choices, taking into account a number of issues whose classification was partially inspired by the list of topics from the Special Track on the Syntactic Analysis of Non-Canonical Language (SPMRL-SANCL 2014).\footnote{\url{http://www.spmrl.org/sancl-posters2014.html}.} These issues include:

\begin{itemize}
    \item sentential unit of analysis, i.e. whether the relevant unit for syntactic analysis is defined by typical sentence boundaries or other criteria
    \item tokenization, i.e. how complex cases of 
    multi-word tokens on the one hand and separated tokens on the other are treated
    \item domain-specific features, such as hashtags, at-mentions, pictograms and other meta-language tokens.
\end{itemize}

\noindent The information on how such phenomena have been dealt with was gathered mostly from the reference papers cited in Table \ref{tab:overview}, and, whenever possible, by searching for the given phenomena within the resources themselves. 

\subsubsection{Sentential unit of analysis}

Sentence segmentation in written text from traditional sources such as newspapers, books or scientific articles is usually defined by the authors through the use of punctuation. However, this is frequently not the case with UGC on social media.\footnote{This is not to say that there is no conventional, well-punctuated data on social media, or that sentence segmentation for other domains is trivial. For instance, many corporations and institutions employ social media managers who adhere to common editing standards. Conversely, some sentence boundaries in canonical written language are also ambiguous, e.g. in headings, tables and captions.} Often, punctuation marks may be missing, misapplied relative to the norms of written language, or used for other communicative needs altogether (e.g. emoticons such as :-|, or emoticons simultaneously serving as closing brackets, etc.). In some cases, no punctuation is used whatsoever, as in Example \ref{ex:nopunct} (the non-standard translation and spelling approximates the lack of punctuation in the original German text).

\begin{examples}
\item \label{exnopunct}Haben
Menschen eigentlich nichts besseres zu tun als Suzie Grime zu haten ja einige Aktionen sind ehrenlos ich habs verstanden \newline 
\textit{Don't people have anything better to do than to hate on Suzie Grime yes some things people do are a disgrace I gettit} \label{ex:nopunct}
\end{examples}

Against this background, it is a non-trivial task to segment social media text manually, let alone automatically. Given that many social media posts by private users tend to consist of sequences of short phrases, clauses and fragments, it is understandable that many Twitter resources consider the entire tweet as a basic unit -- though for other, longer sources, such as Reddit, using entire posts as utterances by analogy is not feasible. Further, certain types of annotations make retaining tweets as single segments more conducive. For instance, \D analyzed the syntactic/semantic relationships and ironic triggers across different sentences, which was more practical with tweets kept intact. In addition, annotation of inter-sentential code-switching (see Section~\ref{sec:UR}) can be considered more appropriate at the tweet level.
Finally, keeping tweets as single units in some treebanks saves the effort needed to develop, maintain, adapt or do post-processing on an automatic sentence segmenter.\footnote{A segmenter could nevertheless be necessary e.g. if the next step is using a parser trained on sentence-split data.}

On the other hand, there are counterbalancing considerations that motivate performing medium-independent segmentation on UGC data, among these a possible overuse of syntactic relations that define side-by-side (or run-on) sentences (e.g. \texttt{parataxis} in UD); second,  as mentioned previously, at least for some UGC data collections (e.g. blog posts), punctuation is found frequently enough and can be used. Third, given that Twitter doubled its character limit for posts from 140 to 280 at the end of 2017, treating tweets as single utterances might pose a usability problem for manual annotation. Finally, for NLP tools trained on multiple genres and for transfer learning, inconsistent sentence spans are likely to reduce segmentation and parsing accuracy. 

Due to these considerations, \rM manually segmented tweets into sentences while introducing an ID system that enables reconstruction of complete posts, if needed. Similarly, \rS uses syntactic utterance level annotations of user IDs and addressee IDs to indicate the post-tree structure in Reddit forum posts. The CoNLL-U format used in the UD project provides the means to implement these kinds of solutions in a straightforward manner, using utterance level comment annotations, which are serialized together with each syntax tree.
\rM, however, still features the use of the \texttt{parataxis} relation within a single utterance for juxtaposed clauses that are not separated by punctuation, even when they form multiple complete sentences, similar to the analysis one would find in newspaper treebanks. 


For other cases authors have introduced additional conventions to cover special constructs occurring in social media. For instance, in some treebanks (sequences of) hashtags and URLs are separated out into `sentences' of their own whenever they occur at the beginning or end of a tweet and do not have any syntactic function.


A third option besides not segmenting and segmenting manually is, of course, to segment automatically.  In the spirit of maintaining a real-world scenario, \rG split their forum data into sentences using NLTK \citep{bird-2004-nltk}, with no post-corrections. Accordingly, the resource contains instances where multiple grammatical sentences are merged into one sentence due to punctuation errors such as a comma being used instead of a full stop, as in Example~ \ref{ex:unit1}. Conversely, there are cases where a single sentence is split over multiple lines, resulting in multiple sentences (Example \ref{ex:unit2}) that are not rejoined.

\vspace{-0.1cm}
\begin{examples}
\item Combofix will start, When it is scanning don't move the
mouse cursor inside the box, can cause freezing.\hfill  (from Foreebank)\label{ex:unit1}
\item I'm sure the devs. \\
can give you more details on this \hfill  (from Foreebank)\label{ex:unit2}
\end{examples}
\vspace{-0.1cm}

\subsubsection{Tokenization}

Tokenization problems in informal text include a wide range of cases which can sometimes require a non-trivial mapping effort to identify the correspondence between syntactic words and tokens. We may thus find multiple words that are merged into a single token, as in contractions\footnote{In this context we take into consideration only the cases encountered in informal/noisy texts, and not the traditional contractions typically present even in each standard language (such as English `don't', the preposition-article contractions in French and German, or the verb-clitic contractions in Italian and German).} (Example \ref{ex:mwt1}, which is also frequent in spoken English and can also be found in literary texts but not in newswire or academic writing) and initialisms such as the Italian example in (\ref{ex:mwt2}), or, conversely, a single syntactic word split up into more than one token (\ref{ex:split1}--\ref{ex:split2} below). 

\begin{examples}
\item gonna $\leftrightarrow$ going to \label{ex:mwt1}
\item \textit{tvtb $\leftrightarrow$ ti voglio tanto bene}\\
I love you so much \label{ex:mwt2}
\end{examples}

We observed a number of different tokenization strategies adopted to deal with those cases but most of the time the preferred solution seemed to involve their decomposition (\rI, \rK, \rM, \rO, \rP,\footnote{In \rI and \rP, however, some examples of phrasal contractions have been found that were not decomposed.} \rS), although a few inconsistencies are found in the resulting lemmatization. Consider  the contraction in Example \ref{ex:mwt1}. \rI  reproduces the same lemma as the word form for both tokens (\textit{gonna$\rightarrow$gon na}), while \rP and \rS instead use its normalized counterpart (\textit{gonna$\rightarrow$go to}).

Alternatively, these contractions might be either decomposed and also normalized by mapping their components onto their standard form, i.e. using `go' and `to' as the normalized word forms and lemmas of a multi-token unit\footnote{These units, sometimes called super-tokens, have a special representation grouping several underlying tokens in the CoNLL-U format and are used to represent phenomena such as preposition-article fusion and other contractions.} `gonna' (\E, \rL,\footnote{In \rL, however, institutionalized and formal abbreviations are not expanded.} \rT), or rather left completely unsplit as a single token (and lemma) `gonna' (\C, \D, \rH, \rN).

\begin{figure}[!ht]
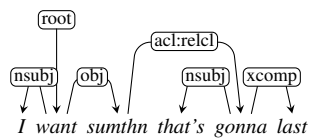

\begin{footnotesize}
\begin{center}
\begin{dependency}
\begin{deptext}
\textit{I} \&[.0001cm] \textit{want} \&[.0001cm] \textit{sumthn}  \&[.0001cm]  \textit{that's} \&[.0001cm] \textit{gonna} \&[.0001cm] \textit{last} \&[.0001cm]\\
\end{deptext}
\deproot[edge unit distance=2.8ex]{2}{\normalsize root}
\depedge{2}{1}{\normalsize nsubj}
\depedge{2}{3}{\normalsize obj}
\depedge{3}{5}{\normalsize acl:relcl}
\depedge{5}{4}{\normalsize nsubj}
\depedge{5}{6}{\normalsize xcomp}
\end{dependency}
\end{center}
\end{footnotesize}
\caption{Example of unsplit contraction from the TAAE treebank.}
\label{fig:tokenize}
\end{figure}

How these cases are annotated syntactically is not always specified in the respective papers, but the general principle seems to be that when contractions are split, the annotation is based on the normalized tokenization (\rI, \rK, \rL, \rO, \rP, \rS), while when they are left unsplit,  annotation is according to the edges connecting words within the phrase's subgraph (\C, \rN). According to this principle, Example \ref{ex:mwt1} would thus be annotated according to the main role played by the verb `go', as shown in Figure~\ref{fig:tokenize}.

As stated above, acronyms and intialisms may also pose a problem for tokenization, but in this case, there seems to be a higher consensus in not splitting them up into individual components, especially where an acronym is established and can be assigned a grammatical function without splitting, e.g. `TL;DR' (too long; didn't read) is left as a single token in \rS, with the reasoning that the form is conventional and likely to be pronounced as the acronym even when read aloud.

When the opposite strategy is used, that of multi-token units, the preferable option, in most cases, is not to merge the separate tokens (\C, \D, \rG, \rI, \rN, \rO, \rP). As a result, one token -- either the first (\C, \D, \rG, \rI, \rN, \rP, \rS) or the last one (\rO) -- is often promoted to represent the main element of the multi-token unit. This kind of ``promotion'' strategy, when put into practice, could actually mean very different things. In \rG, a distinction is drawn between morphological splits (Example \ref{ex:split1}) and simple spelling errors (Example \ref{ex:split2}):

\begin{examples}
\item he should buy \textbf{anti vir} programs $\leftrightarrow$ antivir \hfill  (from Foreebank)\label{ex:split1}
\item \textbf{i t} keeps causing <ProductName> to lock up ...$\leftrightarrow$ it \hfill  (from Foreebank)\label{ex:split2}
\end{examples}

In the first case, both tokens are tagged based on the corresponding category of the intended word, i.e. as a NOUN (since `antivirus' is a noun), while in the second one the two tokens are treated as a spelling error and an extraneous token, respectively. 

In the remaining resources, neither explicit information nor regular/consistent patterns have been found concerning the morpho-syntactic treatment of these units. For their syntactic annotation in dependency grammar frameworks, common practice is to attach all remaining tokens to the one that has been promoted to head status. In UD corpora, the second (and subsequent) tokens in such instances are connected to the first token, and labeled with the special \texttt{goeswith} relation, which indicates superfluous whitespace between parts of an otherwise single token word.

Finally, a distinctive tokenization strategy is adopted in \A with respect to at-mentions, in which the `@' symbol is always split apart from the username, whereas other corpora retain the unsplit username along with the `@' symbol.

\subsubsection{Other domain-specific issues}
This category includes phenomena typical for social media text in general and for Twitter in particular, given that many of the treebanks in this overview contain tweets. Examples include hashtags, at-mentions, emoticons and emojis, retweet markers and URLs.  
These items operate on a meta-language level and are useful for communicating on a social media platform, e.g. for addressing individual users or for adding a semantic tag to a tweet that helps put short messages into context. On the syntactic level, these tokens are usually not integrated, as illustrated for Italian in Example \ref{ex:meta1}.

\begin{examples}
\item \textit{RT @user mi sono davvero divertito :D }\\ 
 RT @user I really had fun :D  \hfill \twadapt{2013}
 \label{ex:meta1}
\end{examples}

It is, however, also possible for those tokens to fill a syntactic slot in the tweet, as shown in the Turkish example in (\ref{ex:meta3}).

\begin{examples}
\item \textit{\#kahvaltı  zamanı}\\ 
time for \#breakfast   \hfill \twsource{2019}
\label{ex:meta3}
\end{examples}
 
In the different treebanks, we observe a very heterogeneous treatment of these meta-language tokens concerning their morpho-syntactic annotation. 
Hashtags and at-mentions, for example, are sometimes  treated as nouns (\E, \rL), as symbols (\D, \rN), or as elements not classifiable according to existing POS categories, or, more generically, as `other' (\rI).

Some resources adopt different strategies that do not fit into this pattern: in \rM and \rS, for example, at-mentions referring to user names are always considered proper nouns while hashtags are tagged according to their respective part-of-speech. Multi-word hashtags are annotated as `other' in \rM (e.g. {\em \#WirSindHandball} `We are handball'), but as proper nouns in \rS ({\em \#IStandWithAhmed}). In \rI, a different POS tag is assigned to at-mentions when they are used in retweets. 

Similarly to hashtags and at-mentions, links can either be annotated as symbols (\D, \rN), nouns (\rF, \rL, \rO), proper nouns (\rS), or `other' (\rM, \rP).\footnote{In \rP the universal POS tag `X' is used, corresponding to the concept of `other', but a special native tag is also applied concurrently: `ADD' (for address), which can then be used to find URLs in particular.} Emoticons and emojis, on the other hand, are mostly classified as symbols (\D, \rI, \rM, \rN, \rP, \rS), less often as interjections (\E, \rO), and in one case as a punctuation mark sub-type (\rL). Retweet markers (\texttt{RT}) are considered as either nouns (\E, \rN) or `other' (\rI\footnote{Except when they are considered an abbreviation of the verb `retweet', in which case they are annotated accordingly.}). On the syntactic level, these meta-tokens are usually attached to the main predicate, but we also observe other solutions. As stated above, in \rM hashtags and URLs at the beginning or end of a tweet form their own sentential units, while in \rH, they are not included in the syntactic analysis.

Finally, in cases where meta-tokens are syntactically integrated, the recurring practice is to annotate them according to their role (\C, \D, \E, \rI \rM, \rN, \rS). \A is unique in that it does not distinguish between meta-tokens at the beginning or end of the tweet and those that are syntactically integrated in the tweet, but instead always assigns a syntactic function to these tokens.

Based on the practices briefly outlined in this section, in the next section, we define an extended inventory of possible annotation issues, some of which occur in only one or a few resources, and propose a set of tentative guidelines for their proper representation within the UD framework.

\section{Towards a Unified Representation}
\label{sec:UR}

In this section we propose a unified approach to annotating the challenges outlined in Section \ref{ssec:comparison}, along with other phenomena that are often found in user-generated text, such as code-switching and disfluencies. We take into consideration the pros and cons for different annotation choices along with the observations we have made across the languages and data sets covered so far in our study.

\subsection{Sentential unit of analysis}
\label{sec:sent}
\begin{figure}[!ht]
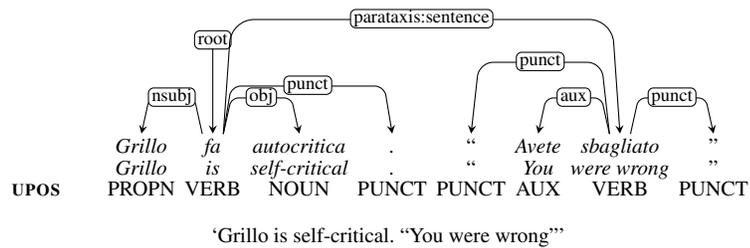

\begin{footnotesize}
\begin{center}
\begin{dependency}
\begin{deptext}
\&[.5cm] \textit{Grillo} \&[.0001cm] \textit{fa} \&[.0001cm] \textit{autocritica}  \&[.0001cm]  \textit{.} \&[.0001cm] `` \&[.0001cm] \textit{Avete} \&[.0001cm] \textit{sbagliato} \&[.0001cm] '' \&[.1cm] \\
\&[.5cm] \textit{Grillo} \&[.0001cm] \textit{is} \&[.0001cm]  \textit{self-critical}   \&[.0001cm]  \textit{.} \&[.0001cm] `` \&[.0001cm] \textit{You} \&[.0001cm] \textit{were wrong} \&[.0001cm] '' \&[.1cm] \\
%
%
\textsc{\textbf{upos}} \&[.5cm] {PROPN} \&[.0001cm]  {VERB} \&[.0001cm] {NOUN} \&[.0001cm] {PUNCT} \&[.0001cm] {PUNCT} \&[.0001cm] {AUX} \&[.0001cm] {VERB}  \&[.0001cm] {PUNCT} \&[.0001cm] 
\\
\end{deptext}
\deproot[edge unit distance=2.8ex]{3}{\normalsize root}
\depedge{3}{2}{\normalsize nsubj}
\depedge{3}{4}{\normalsize obj}
\depedge[edge unit distance=2.5ex]{3}{5}{\normalsize punct}
\depedge[edge unit distance=2.4ex]{3}{8}{\normalsize parataxis:sentence}
\depedge{8}{7}{\normalsize aux}
\depedge{8}{6}{\normalsize punct}
\depedge{8}{9}{\normalsize punct}
\treetrans{`Grillo is self-critical. ``You were wrong'''} 
\end{dependency}
\end{center}
\end{footnotesize}
 \caption{Italian example from Twitter, 2015,  of multiple sentential units in a tweet.}
\label{fig:sentence}
\end{figure}

In the interest of maintaining compatibility with treebanks of standard written language, we propose splitting UGC data into sentential units to the extent to which it is possible and keeping token sequences undivided only when no clear segmentation is possible. 
To facilitate tweet-wise annotation if desired, a subtyped parataxis label, such as \texttt{parataxis:sentence} in Figure~\ref{fig:sentence}, could be used temporarily during annotation. Since some relation label will be needed to connect multiple sentential units within a tweet no matter what, this recommendation is mainly meant to help with later processing or comparison with other data sets, serving as a pointer to identify where the tweet could be split into sentences and distinguishing such junctures from other types of parataxis. 




\subsection{Tokenization}

As shown in the examples in Table \ref{tab:UGCex}, user-generated content can include a number of lexical and orthographic variants whose presence have repercussions with respect to segmentation and choices presented to annotators. The basic principle adopted in UD, for which morphological and syntactic annotation is only defined at the word level \citep{ud:tokenization}, can sometimes clash with the complexity of these cases, 
which has also been a matter of debate within the UD community.\footnote{\url{https://github.com/UniversalDependencies/docs/issues/641}.} \\
 
 \begin{itemize}
     \item \textbf{Contractions}: One particularly challenging issue for annotation decisions related to tokenization is contraction, i.e. when multiple linguistic tokens are contracted to form a single orthographic token (or into fewer tokens than the linguistic content would suggest). It is important to note the different types of contractions that can appear in UGC. For the cases of (i) conventionalized contractions, such as \textit{don't} and (ii) erroneously merged words (e.g \textit{mergedwords}), it is usually easy to identify the morpheme boundary split point. In these cases, we recommend that annotators split the contraction into its component tokens, in keeping with the UD guidelines \citep{ud:boundaryShift} already in place to deal with occurrences of such merging in standard text.

However, for instances of (iii) deliberate informal contractions, such as colloquial abbreviations and initialisms (e.g. EN \textit{gonna, wanna, idk (I don't know)}) or shorthand forms (FR \textit{nimp} `whatever'), standardized criteria are mostly inadequate, or at least insufficient to cover the whole host of possible phenomena. This is due to the ever-changing and often ambiguous nature of user-generated text, i.e. many of the colloquialisms common in UGC are also increasingly conventionalized in the standard language (e.g. {\em gonna}, which is frequent in print in certain registers, and ubiquitous in spoken language), while others may fall out of use entirely. Thus, whether or not a term is considered a conventional contraction is dependent on the time of annotation, and can also be largely subjective. It is also worth noting that increased annotator effort is required if informal contractions are split, as further challenges may be introduced with regard to lemmatization and capturing information for other downstream tasks. This can create a significant overhead in treebank development. For this reason, we advise annotators to adopt an individual approach that takes both treebanking consistency and feasibility into account.

Annotators may wish to consider whether an informal contraction has reached a non-compositional status (e.g. TL;DR, LOL, WTF, idk, etc. in English), and whether it functions solely as a discourse marker or actually bears a semantic and syntactic role within the sentence which is equivalent to its potential expansion (for example, TL;DR, which means`too long; didn't read', is often used in online content creation to provide readers with a shortened summary version of a text). In cases where decomposition of a conventionalized expression is avoided, but the whole function of the phrase is equivalent, our suggested approach is in line with the principle proposed in \newcite{Blodgett2018} where annotation 
is carried out according to the root of the subtree of the original phrase. In the example below, the conventionalized form \textit{idk} (sometimes spelled out when read aloud) is actually used in the place of a matrix verb and is therefore labeled as \texttt{root}, taking a complement clause argument \texttt{ccomp}.



\begin{figure}[!ht]
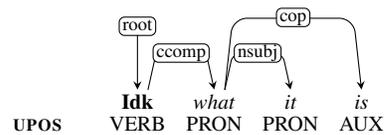

\begin{center} 
\begin{footnotesize}
\begin{dependency}
\begin{deptext}
\&[.5cm] \textbf{Idk} \&[.15cm] \textit{what} \&[.15cm] \textit{it} \&[.15cm] \textit{is} \&[.15cm] \\
\textsc{\textbf{upos}} \&[.5cm] VERB \&[.15cm]  PRON \&[.15cm] PRON \&[.15cm] AUX \&[.15cm] \\
\end{deptext}
\deproot[edge unit distance=2ex]{2}{\normalsize root}
\depedge{2}{3}{\normalsize ccomp}
\depedge{3}{4}{\normalsize nsubj}
\depedge{3}{5}{\normalsize cop}
\end{dependency}
\end{footnotesize}
\end{center}\label{fig:jxt2}
 \caption{Example of an unsplit contraction on Twitter, 2020.}
\end{figure}


Some advantages of leaving deliberate, informal contractions unsplit are that less annotation effort would be required, consistency within the treebank would be easier to maintain, and fewer decisions would be left to the discretion of the annotator (such as the intention of the user and the compositionality of the term in specific instances). Additionally, treebank developers may consider  this approach to be a more descriptive rather than prescriptive representation of `noise' in the data.

By contrast, the benefits of splitting such tokens are that it can be considered a cleaner approach as it will result in fewer ambiguous tokens and it will also allow for more fine-grained detail in the annotation, as well as comparability with resources in which equivalent split forms appear.

\item \textbf{Unconventional Use of Punctuation}: We recommend that unconventional use of punctuation in the form of pictograms {\em :-)} or strings of repeated punctuation marks {\em !!!!!!!} be annotated as a single token rather than being split. Further, we suggest that strings of emoticons be split so that each individual emoticon is considered an individual token, such as {\em :):)} $\rightarrow$ {\em :) + :)} (similar to other sequences of tokens spelled without intervening spaces).

\item \textbf{Over-splitting}: Another tokenization issue relates to the treatment of incorrectly split words. The UD guidelines already advise the use of the \texttt{goeswith} relation in cases of erroneously split words from badly edited texts (e.g. {\em be tween}, {\em gele bilirim}). This means that the split tokens are not merged, but information on their full form is captured nonetheless, while tokens containing whitespace are avoided. In line with the specifications for erroneously split words \citep{ud:boundaryShift} -- be it due to formatting, a typo or intentional splitting -- we suggest to promote the first part of the word to the role of syntactic head and apply left-right attachment, regardless of any potential morphological analysis (i.e. the head of `{\em be tween}' is `{\em be}'). The initial token would also bear the lemma, the POS tag and the morphological features of the entire word, while the remaining split parts would only be POS-tagged as \texttt{X}, and leaving the lemma and features unspecified (by convention `\_'). For instance in the Turkish example in Figure~\ref{fig:oversplit}, \texttt{Number} and \texttt{Person} features, as well as others, are expressed in the \textit{bilirim} part of the over-split word, but annotated in the \textsc{Feats} column of the first part.

\end{itemize}

\begin{figure}[!ht]
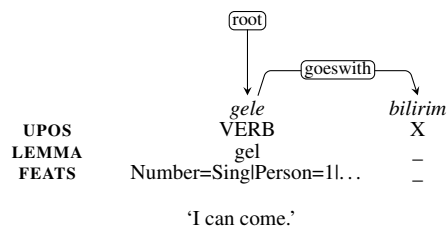

\begin{center}
\begin{footnotesize}
\begin{dependency}
\begin{deptext}
\&[.5cm] \textit{gele} \&[.2cm] \textit{bilirim} \&[.2cm] \\
\textsc{\textbf{upos}} \&[.5cm]     VERB     \&[.2cm]   X    \&[.2cm]  \\
\textsc{\textbf{lemma}}  \&[.5cm]      gel    \&[.2cm]   \_ \&[.2cm]  \\  
\textsc{\textbf{feats}} \&[.5cm]    Number=Sing|Person=1|\dots      \&[.2cm]   \_    \&[.2cm] \\
\end{deptext}
\deproot[edge unit distance=2.3ex]{2}{\normalsize root}
\depedge[edge unit distance=2.6ex]{2}{3}{\normalsize goeswith}
\treetrans{`I can come.'} 
\end{dependency}
\end{footnotesize}
\end{center} 
\caption{Turkish example of over-splitting from Twitter, 2020.}
\label{fig:oversplit}
\end{figure}

\vspace{-0.5cm}


\subsection{Lemmatization}
With respect to the lemmatization of user-generated text, we note that the UD guidelines, specifically those referring to morphology \citep{ud:lemmatization} can often be applied in a straightforward manner. However, certain phenomena common to UGC can complicate this task. In the cases of contraction, over-splitting and unconventional punctuation, lemmatization will depend on the tokenization approach chosen as discussed in the previous section. 

Unconventional uses of punctuation include punctuation reduplication, seemingly random strings of punctuation marks and pictograms or emoticons created using punctuation marks. Punctuation reduplication can be lemmatized by normalizing where a pattern is observed ({\em ?!?!?} $\rightarrow$ {\em ?!}), otherwise the lemma should match the surface form (e.g. {\em !!!!!1!!1} $\rightarrow$ {\em !!!!!1!!1}).
We also recommend that emoticons and pictograms not be normalized ({\em :]]}$\rightarrow$ {\em :]]}), seeing as any attempt of defining a finite set of `conventional' emoticon lemmas would result in a somewhat arbitrary and incomplete list. When lemmatizing neologisms or non-standard vocabulary such as transliterations, we recommend that any inflection be removed in the lemma column (TR {\em taymlaynda} $\rightarrow$ {\em taymlayn}, `(in) timeline'). If the token is uninflected, we suggest the lemma retain the surface form without any normalization (IT {\em tuittare} $\rightarrow$ {\em tuittare}, `to tweet').

\subsection{Features Column}
UD prescribes the use of the features column to list information about the morphological features of the surface form of a word. We suggest that the feature \texttt{Abbr=Yes} be used for abbreviations such as acronyms, initialisms, character omissions, and contractions. Annotators may also choose to include the feature \texttt{Style=X}, employed by some 
 treebanks to describe various aspects of linguistic style such as [\texttt{Coll}: colloquial, \texttt{Expr}: expressive, \texttt{Vrnc}: vernacular, \texttt{Slng}: slang]. Among UGC UD treebanks, only TDT currently uses this feature.

Another useful feature prescribed by UD is \texttt{Typo=Yes} for seemingly accidental deviations from conventional spelling or grammar (used e.g. in \rS, \rP).
The feature \texttt{Foreign=Yes} will be further discussed in Section ~\ref{sec:CS} on code-switching.

\subsection{MISC Column}
At present, aside from capturing instances of spelling variations arising from abbreviation and typos, UD prescribes no mechanism for describing the \emph{nature} of spelling variations. For this reason, we suggest the addition of a new attribute to the UD scheme to denote the more general case of non-canonical language and to more accurately describe the nature of the aforementioned phenomena. This additional attribute \texttt{NonCan=X} would be annotated in the MISC column with the following possible values: 

 [\texttt{AutoC}: autocorrection, \texttt{CharOm}: character omission, \texttt{Cont}: contraction, \texttt{Neo}: neologism, \texttt{OS}: over-splitting, \texttt{Phon}: phonetization, \texttt{PuncVar}: punctuation variation, \texttt{SpellVar}: spelling variation, \texttt{Stretch}: graphemic stretching, \texttt{Transl}: transliteration, \texttt{Trunc}: truncation].\footnote{Examples of each of these are provided in Table~\ref{tab:UGCex}.}

Additionally, the MISC column may be used to list values corresponding to a hypothetical standard or full form of the word,  i.e. the attributes \texttt{CorrectForm=X}, \texttt{FullForm=X}, \texttt{CorrectSpaceAfter=Yes}  may be useful in the cases of non-canonical language, abbreviations and incorrectly merged words respectively.\footnote{At the same time we acknowledge a long strand of research on formulating target hypotheses for non-native and other forms of non-canonical language, which shows that establishing the `correct' or intended form is often a matter of debate requiring detailed guidelines for doubtful cases. See \citet{ReznicekLuedelingHirschmann2013} for discussion.}

The attribute \texttt{LangID=X} will be further discussed in Section ~\ref{sec:CS} on code-switching.

\subsection{Domain-specific issues}
UGC includes many words and symbols with domain-specific meanings. We recommend treating the various  groups as follows:

\begin{itemize}
        \item \textbf{Hashtags} are to be labeled with the X tag as their Universal POS tag (UPOS). Information about the POS category corresponding to the morpho-syntactic role of the token is often captured in the XPOS column based on language specific guidelines e.g. in English, \textit{\#besties}/X/NNS, where NNS designates a plural noun. For languages in which XPOS guidelines do not express the desired distinction, it is possible to use an attribute \texttt{FuncPOS} in the MISC column to express this information. If a hashtag comprises multiple words, it should be kept untokenized, e.g., \textit{\#behappy}/X. Syntactically integrated hashtags should bear their standard dependencies. Classificatory hashtags at the end of tweets are to be attached to the root with the dependency subtype \texttt{parataxis:hashtag} as per the English example in Figure~\ref{fig:hashtags}. 
        
\begin{figure}[!ht]
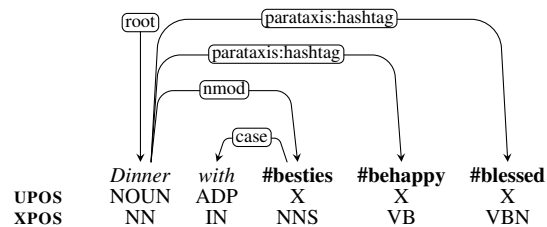

\begin{center}
\vspace{0.5cm}
\begin{footnotesize}
\begin{dependency}
\begin{deptext}
\&[.5cm] \textit{Dinner} \&[.2cm] \textit{with} \&[.2cm] \textbf{\#besties} \&[.2cm] \textbf{\#behappy} \&[.2cm] \textbf{\#blessed}  \\
\textsc{\textbf{upos}} \&[.5cm] NOUN \&[.2cm]  ADP \&[.2cm] X \&[.2cm] X \&[.2cm] X  \\
\textsc{\textbf{xpos}} \&[.5cm] NN \&[.2cm] IN  \&[.2cm] NNS \&[.2cm] VB \&[.2cm] VBN  \\
\end{deptext}
\deproot[edge unit distance=4ex]{2}{\normalsize root}
\depedge[edge unit distance=2.5ex]{4}{3}{\normalsize case}
\depedge{2}{4}{\normalsize nmod}
\depedge{2}{5}{\normalsize parataxis:hashtag}
\depedge{2}{6}{\normalsize parataxis:hashtag}
\end{dependency}
\end{footnotesize}
\vspace{-0.1cm}
\end{center}
 \caption{English example of hashtag usage from Twitter, 2018.}
 \label{fig:hashtags}
\end{figure}


\item \textbf{At-mentions} to be labelled as PROPN. Their syntactic treatment is similar to hashtags: when in context they bear the actual syntactic role (see Figure~\ref{fig:at} for a Turkish example), otherwise they should be dependent on the main predicate with the \texttt{vocative} label as per the Irish example in Figure~\ref{fig:at2}. 

\begin{figure}[!ht]
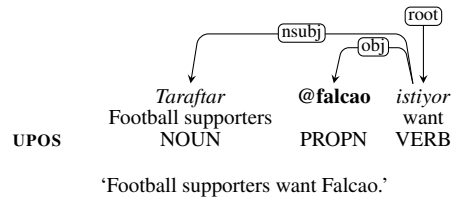

\begin{footnotesize}
\begin{center}
\vspace{0.5cm}
\begin{dependency}
\begin{deptext}
\&[.5cm] \textit{Taraftar} \&[.2cm] \textbf{@falcao} \&[.2cm] \textit{istiyor}  \&[.2cm] \\
\&[.5cm] Football supporters \&[.2cm] \&[.2cm] want  \&[.2cm] \\
\textsc{\textbf{upos}} \&[.5cm]  NOUN \&[.2cm] PROPN \&[.2cm] VERB   \&[.2cm]  \\
\end{deptext}
\depedge[edge unit distance=2.6ex]{4}{2}{\normalsize nsubj}
\depedge{4}{3}{\normalsize obj}
\deproot[edge unit distance=2.1ex]{4}{\normalsize root}
\treetrans{`Football supporters want Falcao.'} 
\end{dependency}
\end{center}
\vspace{-0.1cm}
\end{footnotesize}
 \caption{Turkish example of a syntactically incorporated at-mention from Twitter, 2018.}
 \label{fig:at}
\end{figure}


\begin{figure}[h!]
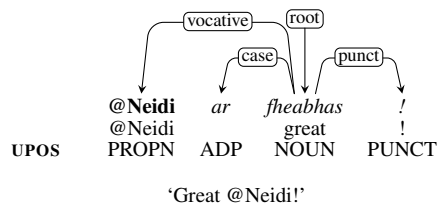

\begin{footnotesize}
\vspace{0.5cm}
\begin{center}
\begin{dependency}
\begin{deptext}
\&[.5cm] \textbf{@Neidi} \&[.2cm] \textit{ar} \&[.2cm] \textit{fheabhas} \&[.2cm] \textit{!}  \&[.2cm] \\
\&[.5cm] @Neidi \&[.2cm] \&[.2cm] great \&[.2cm] !  \&[.2cm]\\
\textsc{\textbf{upos}} \&[.5cm]  PROPN \&[.2cm] ADP \&[.2cm] NOUN \&[.2cm] PUNCT  \&[.2cm]  \\
\end{deptext}
\depedge{4}{2}{\normalsize vocative}
\depedge{4}{3}{\normalsize case}
\deproot[edge unit distance=2.3ex]{4}{\normalsize root}
\depedge{4}{5}{\normalsize punct}
\treetrans{`Great @Neidi!'} 
\end{dependency}
\end{center}
\end{footnotesize}
 \caption{Irish example of a vocative at-mention from Twitter, 2012.}
 \label{fig:at2}
\end{figure}


\item \textbf{URLs} are to be tagged as SYM as per UD guidelines. They are often appended at the end of the tweet without bearing any syntactic function. Throughout our explored corpora, those URLs are diversely annotated, without an obvious consensus emerging: \texttt{parataxis:url} vs \texttt{discourse:context} vs \texttt{dep}. In cases where they are syntactically integrated in the sentence, we recommend that the XPOS should be recorded as NOUN if there are no existing XPOS tags for the language, otherwise the regular XPOS tagging guidelines apply. The token can be attached to its head with the appropriate dependency label as per Figure~\ref{fig:url}. We favour using \texttt{parataxis:url} for non-syntactically integrated URLs.

\begin{figure}[h!]
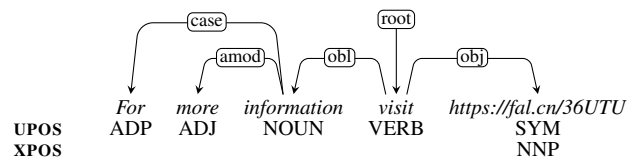

\begin{center}
\vspace{0.5cm}
\begin{footnotesize}
\begin{dependency}
\begin{deptext}
\&[.5cm] \textit{For} \&[.2cm] \textit{more} \&[.2cm] \textit{information} \&[.2cm] \textit{visit} \&[.2cm] \textit{https://fal.cn/36UTU}\\
\textsc{\textbf{upos}}\&[.5cm]  ADP \&[.2cm] ADJ \&[.2cm] NOUN \&[.2cm] VERB  \&[.2cm] SYM\\
\textsc{\textbf{xpos}} \&[.5cm]   \&[.2cm]  \&[.2cm]  \&[.2cm]  \&[.2cm] NNP \\
\end{deptext}
\depedge{4}{2}{\normalsize case}
\depedge{4}{3}{\normalsize amod}
\depedge{5}{4}{\normalsize obl}
\deproot[edge unit distance=2.3ex]{5}{\normalsize root}
\depedge{5}{6}{\normalsize obj}
\end{dependency}
\end{footnotesize}
\vspace{-0.1cm}
\end{center}
 \caption{English example of URL attachment, from Twitter, 2020.}
 \label{fig:url}
\end{figure}


        \item \textbf{Pictograms} are often used at the end of the tweets as discourse markers. In such cases they should be POS-tagged as SYM and attached to the root with the \texttt{discourse} relation. But there are also cases where they replace an actual word (or morpheme) in a syntactic context and  may also be inflected (cf. \ref{ex:pastt}--\ref{ex:ppl}).

\vspace{0.3cm}
\begin{examples}
\item \includegraphics[height=8px]{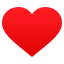}ed it \hfill \twsource{2020}
\label{ex:pastt}

\item Thank you 4 All U do \& \includegraphics[height=8px]{heart.png}ing dogs \hfill \twsource{2020} \label{ex:ppl}
\end{examples}
\vspace{0.3cm}
        
These cases  are to be annotated with the lemma, UPOS tag and dependency relation of the word they substitute. The French example in Figure~\ref{fig:picto} demonstrates both cases.

\begin{figure}[h!]
\begin{center}
\vspace{0.5cm}
\begin{footnotesize}
\begin{dependency}
\begin{deptext}
\&[.5cm] \textit{j'} \&[.2cm] \includegraphics[height=8px]{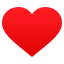} \&[.2cm] \textit{café} \&[.2cm] \Coffeecup\\
\&[.5cm] \textit{I} \&[.2cm] love  \&[.2cm] \textit{coffee} \&[.2cm] \\
\textsc{\textbf{lemma}} \&[.5cm]  je \&[.2cm] aimer  \&[.2cm] café \&[.2cm] SYM  \\
\textsc{\textbf{upos}} \&[.5cm]  PRON \&[.2cm] VERB \&[.2cm] NOUN \&[.2cm] SYM  \\
\textsc{\textbf{xpos}} \&[.5cm] PRO:PER  \&[.2cm] VER:pres \&[.2cm] NOM  \&[.2cm] SYM  \\
\textsc{\textbf{feats}} \&[.5cm]   \&[.2cm] Tense=Present|\ldots \&[.2cm]  \&[.2cm]   \\
\end{deptext}
\depedge{3}{2}{\normalsize nsubj}
\deproot[edge unit distance=2.3ex]{3}{\normalsize root}
\depedge{3}{4}{\normalsize obj}
\depedge{3}{5}{\normalsize discourse}
\end{dependency}
\end{footnotesize}
\vspace{-0.1cm}
\end{center}
 \caption{French example of differing syntactic roles of pictograms from Twitter, 2013.}
 \label{fig:picto}
\end{figure}

The morphological features should reflect the intended meaning. Thus, in example \ref{ex:misinfl} the feature for the pictogram/verb should be \texttt{Person=3} even though the form canonically is a non-third-person form.

\vspace{0.3cm}
\begin{examples}
\item Go follow @IMAPCT$\_$Zodiak he’s a beast \& he'll follow back. He \includegraphics[height=8px]{heart.png} his followers.
\hfill \twsource{2020}\label{ex:misinfl} 
\end{examples}
\vspace{0.3cm}


\item \textbf{RT}s are originally used with at-mentions so that the Twitter interface interprets it as a retweet, as in Figure \ref{fig:retweet1}. In such cases, their UPOS should be SYM with a dependency label \texttt{parataxis} attached to the root. However they are now more commonly used as an abbreviation for \textit{retweet} within a tweet. The UPOS tag should be NOUN or VERB depending on its syntactic role. In these cases, the dependency relation depends on the functional role of the full form (see Figure~\ref{fig:retweet2}).
            
\begin{figure}[h!]  
\begin{center}
\vspace{0.5cm}
\begin{footnotesize}
\begin{dependency}
\begin{deptext}
\&[.5cm] \textit{RT} \&[.2cm] \textit{@user} \&[.2cm] \textit{mi} \&[.2cm] \textit{sono} \&[.2cm] \textit{davvero} \&[.2cm] \textit{divertito} \&[.2cm] \textit{:D}\\
\&[.5cm] \&[.2cm]  \&[.2cm] \textit{I} \&[.2cm]   \&[.2cm] \textit{really} \&[.2cm] \textit{had fun} \&[.2cm] \\
\textsc{\textbf{upos}} \&[.5cm]  SYM \&[.2cm] PROPN \&[.2cm] PRON \&[.2cm] VERB \&[.2cm] ADV \&[.2cm] ADJ  \&[.2cm] SYM\\
\end{deptext}
\depedge{7}{2}{\normalsize parataxis}
\depedge{7}{3}{\normalsize vocative}
\depedge{7}{4}{\normalsize expl}
\depedge{7}{5}{\normalsize cop}
\depedge{7}{6}{\normalsize advmod}
\deproot[edge unit distance=4.1ex]{7}{\normalsize root}
\depedge{7}{8}{\normalsize discourse}
\treetrans{`I really had fun'} 
\end{dependency}
\end{footnotesize}
\vspace{-0.1cm}
\end{center}
 \caption{Italian example of usage of `RT' from Twitter, 2013}
 \label{fig:retweet1}
\end{figure}

\begin{figure}[h!]
\begin{center}
\vspace{0.5cm}
\begin{footnotesize}
\begin{dependency}
\begin{deptext}
\&[.5cm] \textit{Bitte} \&[.2cm] \textit{diesen} \&[.2cm] \textit{Beitrag} \&[.2cm] \textbf{RT}\\
\&[.5cm] Please \&[.2cm] this \&[.2cm] article \&[.2cm] \textbf{retweet}\\
\textsc{\textbf{upos}} \&[.5cm]  INTJ \&[.2cm] DET \&[.2cm] NOUN \&[.2cm] VERB  \\
\end{deptext}
\depedge[edge unit distance=2.3ex]{5}{2}{\normalsize discourse}
\depedge{4}{3}{\normalsize det}
\depedge{5}{4}{\normalsize obj}
\deproot[edge unit distance=2.5ex]{5}{\normalsize root}
\treetrans{`Please retweet this article.'}
\end{dependency}
\end{footnotesize}
\vspace{0.5cm}
\end{center}
 \caption{German example of RT as a verb -- from Twitter, 2019.}
 \label{fig:retweet2}
\end{figure}


        \item \textbf{Markup} symbols (e.g. <, >, +++) have the UPOS SYM -- similar to e.g., math operators in the UD guidelines -- and we recommend they be attached to the head with \texttt{punct} as per the German example in Figure~\ref{fig:markup}.
        
%

\begin{figure}[h!]
\begin{center}
\vspace{-0.1cm}
\begin{footnotesize}
\begin{dependency}
\begin{deptext}
 
\&[.5cm] \textit{+++} \&[.2cm] \textit{Süßigkeiten} \&[.2cm] \textit{nicht} \&[.2cm] \textit{vergessen} \&[.2cm] \textit{!!} \&[.2cm] \textit{+++}\\
\&[.5cm]   \&[.2cm] sweets \&[.2cm] not \&[.2cm] \textit{forget} \&[.2cm] \&[.2cm]\\
\textsc{\textbf{upos}} \&[.5cm]  SYM \&[.2cm] NOUN \&[.2cm] ADV \&[.2cm] VERB \&[.2cm] PUNCT \&[.2cm] SYM\\
\end{deptext}
\depedge[edge unit distance=2.5ex]{5}{2}{\normalsize punct}
\depedge[edge unit distance=2.3ex]{5}{3}{\normalsize obj}
\depedge{5}{4}{\normalsize advmod}
\deproot[edge unit distance=2.9ex]{5}{\normalsize root}
\depedge{5}{6}{\normalsize punct}
\depedge{5}{7}{\normalsize punct}
\treetrans{`Do not forget sweets!'}
\end{dependency}
\end{footnotesize}
\vspace{-0.1cm}
\end{center}
 \caption{German example of the use of markup symbols from Twitter, 2020.}
 \label{fig:markup}
\end{figure}


\end{itemize}

\subsection{Code-switching}
\label{sec:CS}
 As discussed in Section~\ref{sec:web}, capturing code-switching (CS) in tweets is an additional motivation for following a tweet-based unit of analysis \citep{cetinoglu-2016-turkish,lynn-2019}. CS -- switching between languages -- is an emerging topic of interest in NLP \citep{Solorio:2008,Solorio2014,Bhat2018a} and as such should be captured in treebank data where possible.  CS can occur on a number of levels. CS that occurs at the sentence or clause level is referred to as inter-sentential switching (INTER) as shown between English and Irish in Example~\ref{ex:inter} and German and Turkish in Example~\ref{ex:inter2}:

\begin{examples}
\item ``\textit{Má tá AON Gaeilge agat, úsáid í!} It's Irish Language Week.''\\
If you have ANY Irish, use it! It's Irish Language Week. \hfill \twsource{2014}
\label{ex:inter}
\end{examples}


\begin{examples}
\item ``\textit{@user Jedem das was er verdient. ;-) Yoksa Köln'den Almanca ö\u{g}renmeden mi döndün}''\\
Everyone gets what they deserve ;-) Or did you return from Cologne without learning German?
\hfill \twsource{2014}
\label{ex:inter2}
\end{examples}

INTER switching can also be used to describe bilingual tweets where the switched text represents a translation of the previous segment: ``Happy St Patrick’s Day! \textit{La Fhéile Pádraig sona daoibh!''} This phenomenon is often seen in tweets of bi-/multi-lingual users.\\


CS occurring within a clause or phrase is referred to as intra-sentential switching (INTRA). Example~\ref{ex:intra} demonstrates INTRA switching between Italian and English:\\



\begin{examples}
\item ``\textit{Le proposte per l'\textbf{education} di Confindustria}''\\
‘The proposals for the education by Confindustria' \hfill (adapted from TWITTIRÒ, 2014)
\label{ex:intra}
\end{examples}


\vspace{-0.1cm}
Word-level alternation (MIXED) describes the combination of morphemes from different languages or the use of inflection according to rules of one language in a word from another language. This is particularly evident in highly inflected or agglutinative languages. Example~\ref{ex:word} shows the creation of a Turkish verb derived from the German noun \textit{Kopie} `copy'.


\begin{examples}
\item \textit{Adamın 3-4 biyografisi var \textbf{Kopie}lenip yapıştırılmış.}\\
`The guy has 3-4 biographies copied and pasted.' \hfill \twadapt{2016} \\
\label{ex:word}
\end{examples}

While borrowed words can often become adopted into a language
over time (e.g. \textit{cool} is used worldwide), when a word is still regarded as foreign in the context of CS, the suggested UPOS is the switched token's POS -- if known or meaningful -- otherwise X is used \citep{ud:foreignPOS}.
The morphological feature \texttt{Foreign=Yes} should be used, and we also suggest that the language of code-switched text is captured in the MISC column, along with an indication of the CS type. As such, in Example~\ref{ex:intra}, \textit{education} would have the MISC values of \texttt{CSType=INTRA | LangID=EN}.



\begin{figure}[h!]
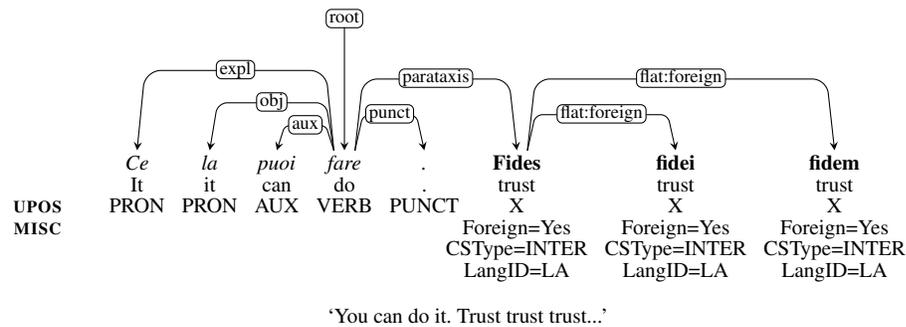

\begin{center}
\begin{footnotesize}
\begin{dependency}
\begin{deptext}
\&[.5cm] \textit{Ce} \&[.1cm] \textit{la} \&[.1cm] \textit{puoi} \&[.1cm] \textit{fare} \&[.1cm] \textit{.} \&[-.3cm] \textbf{Fides} \&[.1cm] \textbf{fidei} \&[.1cm] \textbf{fidem} \&[.1cm] \\ 
\&[.5cm]   It     \&[.1cm]   it    \&[.1cm]  can   \&[.1cm]  do   \&[.1cm]   . \&[-.3cm]   trust \&[.1cm]   trust \&[.1cm]   trust \&[.1cm] \\  
\textsc{\textbf{upos}} \&[.5cm]     PRON     \&[.1cm]   PRON    \&[.1cm]  AUX   \&[.1cm]  VERB   \&[.1cm] PUNCT \&[-.3cm] X  \&[.1cm] X  \&[.1cm] X   \&[.1cm] \\
\textsc{\textbf{misc}} \&[.5cm]         \&[.1cm]    \&[.1cm] \&[.1cm] \&[.1cm]   \&[-.3cm]  Foreign=Yes   \&[.1cm]   Foreign=Yes  \&[.1cm] Foreign=Yes   \\  
\&[.5cm]         \&[.1cm]  \&[.1cm] \&[.1cm] \&[.1cm]      \&[-.3cm]  CSType=INTER   \&[.1cm]    CSType=INTER  \&[.1cm]  CSType=INTER   \\  
\&[.5cm]         \&[.1cm] \&[.1cm] \&[.1cm] \&[.1cm]       \&[-.3cm]  LangID=LA   \&[.1cm] LangID=LA    \&[.1cm] LangID=LA   \\  
\end{deptext}
\depedge[edge unit distance=2.8ex]{5}{2}{\normalsize expl}
\depedge[edge unit distance=2.5ex]{5}{3}{\normalsize obj}
\depedge[edge unit distance=2.6ex]{5}{4}{\normalsize aux}
\deproot[edge unit distance=3.8ex]{5}{\normalsize root}
\depedge{5}{6}{\normalsize punct}
\depedge{5}{7}{\normalsize parataxis}
\depedge{7}{8}{\normalsize flat:foreign}
\depedge{7}{9}{\normalsize flat:foreign}
\treetrans{`You can do it. Trust trust trust...'}
\end{dependency}
\end{footnotesize}
\end{center}
 \caption{Italian-Latin code-switching example tree, adapted from POSTWITA, 2011.}
 \label{fig:CS1}
\end{figure}

\begin{figure}[h!]
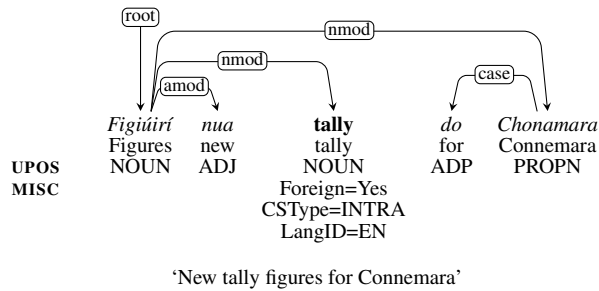

\begin{center}
\begin{footnotesize}
\begin{dependency}
\begin{deptext}
\&[.5cm] \textit{Figiúirí} \&[.2cm] \textit{nua} \&[.2cm] \textbf{tally} \&[.2cm] \textit{do} \&[.2cm] \textit{Chonamara} \\
\&[.5cm]   Figures     \&[.2cm]   new    \&[.2cm]  tally   \&[.2cm]  for   \&[.2cm]   Connemara  \&[.2cm]\\  
\textsc{\textbf{upos}} \&[.5cm]     NOUN     \&[.2cm]   ADJ    \&[.2cm]  NOUN   \&[.2cm]  ADP   \&[.2cm] PROPN   \&[.2cm] \\  
\textsc{\textbf{misc}} \&[.5cm]          \&[.2cm]       \&[.2cm]  Foreign=Yes   \&[.2cm]     \&[.2cm]   \&[.2cm]  \\  
\&[.5cm]         \&[.2cm]       \&[.2cm]  CSType=INTRA   \&[.2cm]     \&[.2cm]   \&[.2cm]  \\  
\&[.5cm]         \&[.2cm]       \&[.2cm]  LangID=EN   \&[.2cm]     \&[.2cm]  \&[.2cm]  \\  
\end{deptext}
\deproot[edge unit distance=2.8ex]{2}{\normalsize root}
\depedge[edge unit distance=2.5ex]{2}{3}{\normalsize amod}
\depedge[edge unit distance=2.6ex]{2}{4}{\normalsize nmod}
\depedge[edge unit distance=2.1ex]{2}{6}{\normalsize nmod}
\depedge{6}{5}{\normalsize case}
\treetrans{`New tally figures for Connemara'}
\end{dependency}
\end{footnotesize}
\end{center}
 \caption{Irish-English code-switching example from Twitter, 2014.}
\label{fig:CS2}
\end{figure}

In terms of syntactic annotation, the UD guideline recommends that the \texttt{flat} or \texttt{flat:foreign} label is used to attach all words in a foreign string to the first token of that string \citep{ud:foreign}. We recommend that this guideline is followed (for both INTER and INTRA CS) when the grammar of the switched text is not known to annotators (see Figure~\ref{fig:CS1}). Otherwise, we recommend applying the appropriate syntactic analysis for the switched language (see Figure~\ref{fig:CS2}).





Lemmatization of CS tokens can prove difficult if a corpus contains multiple languages that annotators may not be familiar with. To enable more accurate cross-lingual studies,
all switched tokens should be (consistently) lemmatized if the language is known to annotators and annotation is feasible within the constraints of a treebank's development phase. Otherwise the surface form should be used, allowing for more comprehensive lemmatization at a later date.


\subsection{Disfluencies}
Similarly to spoken language, UGC often contains disfluencies such as repetitions, fillers or aborted sentences. This might be surprising, given that UGC does not pose the same pressure on cognitive processing that online spoken language production does.



In UGC, however, what may seem to be a production
error can in fact have a completely different function  \citep{rehbein:2015}. Here, repetitions, self-repair and hesitation markers are often used with humorous intent (Example \ref{ex:humor} demonstrates this for German). 

\begin{examples}
\item \textit{Du hast den Apple Wahnsinn... \"ah, Spirit einfach noch nicht verstanden ;)}\\
`You haven’t yet understood the Apple madness... uh spirit ;)' \\ 
{\color{white}{...}} ~ \hfill \twsource{2012}
\label{ex:humor}
\end{examples}


Disfluencies pose a major challenge for syntactic analysis as they often result in an incomplete structure or in a tree where duplicate lexical fillers compete for the same functional slot. 
 
\begin{figure}[h!]
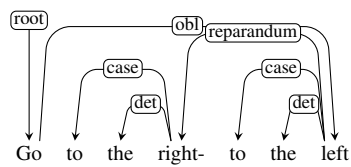

\begin{center}
\begin{footnotesize}
\begin{dependency}
\begin{deptext}
Go \&[.2cm] to \&[.2cm] the \&[.2cm] right- \&[.2cm] to \&[.2cm] the \&[.2cm] left \&[.2cm] \\
\end{deptext}
\deproot[edge unit distance=3.5ex]{1}{\normalsize root} 
\depedge[edge unit distance=2ex]{1}{7}{\normalsize obl}
\depedge{4}{3}{\normalsize det}
\depedge{4}{2}{\normalsize case}
\depedge{7}{6}{\normalsize det}
\depedge{7}{5}{\normalsize case}
\depedge{7}{4}{\normalsize reparandum}
\end{dependency}
\end{footnotesize} \label{ex:reparandum}
\end{center}
 \caption{Example tree from the UD guidelines, showing the use of \texttt{reparandum}.}
 \label{fig:UDdis}
\end{figure}

For UD, some treebanks with spoken language material exist \citep{rhapsodie:2014,dobrovoljc:2016,leung:2016,ovrelid:2016,caron:etal:2019}, and among the treebanks surveyed here, \rS also contains some speech data. The UD guidelines propose the following analysis for disfluency repairs \citep{ud:reparandum} (See Figure~\ref{fig:UDdis}).

This treatment loses information whenever the reparandum does not carry the same grammatical function as the repair, as illustrated in German in Example \ref{ex:stallion}.
In this example from  Twitter, the user plays with the homonymic forms of the German noun {\em Hengst} (stallion) and the verb {\em h\"angst} (hang$_{2.Ps.Sg}$).
The missing information, however, can be easily added when using the enhanced UD scheme \citep{ud:enhanced}.

\begin{examples}
\item Du ~Hengst! ~\"ah, h\"angst.\\
You stallion! uh, hang$_{2.Ps.Sg}$.\\
``You stallion! uh, you're stalled.'' \hfill \twsource{2012} \label{ex:stallion}
\end{examples}

%

Other open questions concern the use of hesitation markers in UGC. We propose to consider them as multi-functional discourse structuring devices and annotate them as discourse markers, attached to the root, a practice already followed for example in \rS for both spoken data and reported speech within UGC.

\section{Discussion}
 In this final section, we discuss some open questions in which the nature of the phenomena described makes their encoding difficult by means of the current UD scheme. 

\subsection{Elliptical structures and missing elements}
\label{sec:ellipsis}
In constituency-based treebanks 
that contain canonical texts, 
such as the Penn Treebank \citep{marcus1993building}, the annotation of empty elements results from the need to keep traces of movement and long-distance dependencies, usually marked with trace tokens and co-indexing at the
lexical level in addition to the actual nodes dominating such empty elements. The dependency syntax framework usually does not use such devices as these syntactic phenomena can be represented with crossing branches resulting in non-projective trees. 

In the specific case of gapping coordination, which can be analyzed as the results of the deletion of a verbal predicate (e.g. John loves$_i$ Mary and Paul (e$_i$) Virginia), both the subject and object of the right-hand side conjunct are annotated with the \texttt{orphan} relation \citep{schuster-etal-2017-gapping}. Even though the Enhanced UD scheme proposes to include a {\em ghost}-token \citep{SchusterManning:2016:EnhancedEU} which will be the actual governor of the right hand-side conjuncts, 
nothing is prescribed regarding the treatment of ellipsis without an antecedent. Given the contextual nature of most UGC sources and their space constraints, those cases are very frequent. The problem lies in the interpretation underlying some annotation scenarios. \newcite{Martinez2016} analyzed an example from a French video game chat log where all verbs were elided. Depending on contextual interpretation of a modifier, a potential analysis could result in two concurrent trees. Such an analysis is not allowed in the current UD scheme, unless the trees are duplicated and one analysis is provided for each of them.

\begin{figure}[!ht]
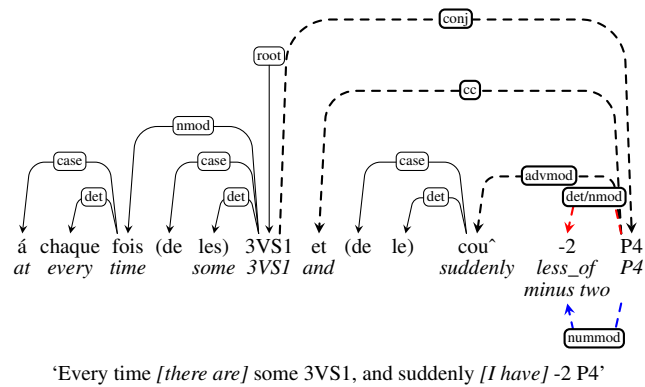

\centering
\small
\begin{dependency}
\begin{deptext}
\'a \& chaque\& fois \&  (de \& les)\&  3VS1\& et\& (de  \& ~~le)~~~ \& cou\^{} \& -2 \&  P4\\
\textit{at} \& \textit{every} \& \textit{time} \& \textit{~} \& \textit{some} \& \textit{3VS1} \& \textit{and} \&  \&  \& \textit{suddenly} \& \textit{less\_of}  \&\textit{P4}\\
\&  \&  \&  \&  \&  \&  \& \& \&  \&  \textit{minus two} \& \\
\end{deptext}
\depedge[edge style={black}]{3}{1}{case}
\depedge[edge style={black}]{3}{2}{det}
\depedge[edge style={black}]{6}{3}{nmod}
\depedge[edge style={black}]{6}{4}{case}
\depedge[edge style={black}]{6}{5}{det}
\deproot[edge unit distance=5ex]{6}{root}
\depedge[style=dashed,black,thick, edge unit distance=3ex]{12}{7}{cc}
\depedge[edge style=dashed, black,thick, edge unit distance=3ex]{12}{10}{advmod}
\depedge[edge style={black,arc edge}]{10}{8}{case}
\depedge[edge style={black,arc edge}]{10}{9}{det}

\depedge[style=dashed,black,thick]{6}{12}{conj}
\depedge[style=dashed,red,thick]{12}{11}{det/nmod}
\depedge[style=thick,dashed,blue,edge below]{12}{11}{nummod}
\treetransbelow{`Every time \textit{[there are]} some 3VS1, and suddenly \textit{[I have]} -2 P4'}
\end{dependency}
\caption{Problematic example with two contesting structures from two different readings of the token ``-2'' surrounded by at least 2 elided elements. \emph{(Adapted to UD v2.5 from \citep{Martinez2016}})
}
\label{fig:example:pathological}
\end{figure}

Following from a French example taken from \citep{Martinez2016}, Figure \ref{fig:example:pathological} shows an attachment ambiguity caused by part-of-speech ambiguity and verb ellipsis. A natural ellipsis recovery of the example shown in Figure \ref{fig:example:pathological} would read as `Every time \textbf{there are} 3VS1, and suddenly \textbf{I have} -2 P4'. The token "3VS1" stands for ``3 versus 1'', namely an uneven combat setting, and `P4'' refers to a {Minecraft} character's protection armor. The token "-2" allows for more than one analysis. The first analysis is the simple reading as number, complementing the noun "P4", in blue in the graph below. A second analysis, in red, treats "-2" as a transcription of \textit{moins de} (less of), which would be the preferred analysis given  the P4 as an armor level interpretation. This example shows the interplay between frequent ellipses, ergographic phenomena and the need for domain knowledge in user-generated data. It also highlights the importance of annotators' choices when facing elided content as it would have been perfectly acceptable to use the {\tt orphan} relation to mark the absence of, in this a case, a verbal predicate (eg. {\tt orphan(3VS1,fois)} and {\tt orphan(p4,cou\^{})}). The next paragraph illustrates such an analysis on a case from German Twitter.    



The German example in Figure~\ref{fig:ellipsis} below illustrates a type of antecedent-less ellipsis that occurs in the spoken commentary of sportscasters but is also used on Twitter by users who mimic such play-by-play commentary on football games they are watching in real time.  As with the French video chat example in Figure~\ref{fig:example:pathological}, it is not clear which verb should be reconstructed in the first elliptical clause as there is no antecedent in the prior discourse. Given the context -- Müller and Lewandowski are well known football players (in Germany) and the preposition \textit{auf} signals a motion event, it is clear that the first conjunct reports an event where one player passes the ball to another. But the specific manner in which the ball is moved, whether it is  `headed' or `kicked', could only be determined by watching footage of the game. 
The example also illustrates a second verb-less clause that is potentially difficult to recognize: `TOR' heads  its own clause and is coordinated with \#Müller rather than being conjoined to \#Lewandowski. The relevant clue is the capitalization that evokes the loudness and emphasis of the goal cheer. Again, one cannot be fully confident which verb to reconstruct here: several existential-type verbs are conceivable.

\begin{figure}[!ht]
\centering
\small
\begin{dependency}
\begin{deptext}
 \#Müller  \& auf  \& \#Lewandowski \& und \& das \& TOR\& !\\
 \#Müller  \& to  \& \#Lewandowski \& and  \& the \& GOAL\& !\\
\end{deptext}
\deproot[edge unit distance=5ex]{1}{root}
\depedge{3}{2}{case}
\depedge{6}{5}{det}
\depedge{6}{4}{cc}
\depedge{1}{3}{orphan}
\depedge{1}{6}{conj}
\treetrans{\#Müller (passes/heads/kicks/...) to Lewandowski and (there is) the GOAL!}
\end{dependency}
\caption{German example of ellipsis from Twitter, 2020.}
\label{fig:ellipsis}
\end{figure}

Example \ref{ex:james} below illustrates a further variation of the above case in German:  the PPs can be iterated to iconically capture a series of passes. Thus, in the example below, Müller is not only the recipient of the ball from James but also the one that passes it on to Lewandowski. However, it is not clear what structure to assume in an enhanced UD analysis that would explicate this. One could assume (i) two explicitly coordinated clauses, (ii) two clauses related by parataxis, or (iii) the use of a relative clause for the second clause. None of these analyses would be obviously right or wrong.

\begin{examples}
\item \label{ex:james}  TOR ~~ - James auf Müller auf \#Lewandowski: die Entscheidung (87.)\\
GOAL - James to ~ Müller to \#Lewandowski: the clincher (87.)\\\textcolor{white}{xxx} \hfill \twsource{2020}
\end{examples}

In any event, it is very likely that the growing number of UD treebanks containing user-generated content (and/or spoken language) will be found to feature many constructions that cannot readily be handled based on the existing guidelines for written language. 


\subsection{Limitations}
In focusing only on user-generated content at the sentence level, our proposal does not cover phenomena that spread over multiple sentences, which would be relevant at a discourse annotation level and seen for example in cases of extra-sentential references. In the case of threaded discussions, similar to dialogue interaction, cases of gapping and more generally syntactic ellipsis can occur. These are not covered by our proposal, nor are they permitted in the UD framework, as they would require a more elaborate token indexing scheme spanning over sentences.  

\begin{figure}[!ht]
\begin{minipage}[b]{.50\textwidth}
\centering
\begin{verbatim}
NN   N  OOOOOO
N N  N  O    O
N  N N  O    O
N   NN  OOOOOO
\end{verbatim}
\caption{ASCII art example (NO), adapted from Twitter, 2018.}\label{fig:ascii}
\end{minipage}
\hfill
\begin{minipage}[b]{.45\textwidth}
\begin{tabular}{lll}
& & \\
Not eating  	&   &   feeling sick bc \\
Bc I feel sick  & \includegraphics[height=8px]{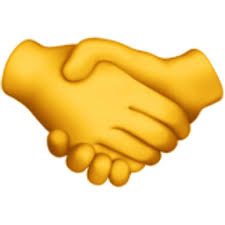}  &	I'm not eating\\
& & \\
& & \\
\end{tabular}
\caption{Two-column tweet example from Twitter, 2020.}\label{fig:twocolumns}
\end{minipage}
\end{figure}

Like any content expressed in digital (i.e. non-handwritten) media, any conceivable variation of {\em ASCII art} can be used and carry meaning (Figure~\ref{fig:ascii}). Formatting variations, such as a recent trend of two-column tweets, as shown in Figure \ref{fig:twocolumns}, are observed where some graphical layout recognition is needed to interpret the two columns as two consecutive sentences. This is similar to challenges in standard text corpora acquired from visual media, such as literary corpora from multi-column pages digitized by Optical Character Recognition (OCR). Cases such as this do not require any specific annotation as this proposal refers to the processing of (mostly) text-based interactions.

Another difficult phenomenon to annotate lies in the multi-modal nature of most user-generated content platforms that enable the inclusion of various media contents (picture, video, etc.) that often provide context to a tweet or provide meta-linguistic information that changes the whole interpretation of this content. While those phenomena do not change the core syntactic annotation {\em per se}, they can change the way that tokens such as proper noun strings or URLs can be interpreted. 


\subsection{Interoperability and other frameworks}

\paragraph{UD enhancements and modifications}
The UD framework is an evolving, ongoing community effort, informed by contributors from wide and varied linguistic backgrounds. 
One line of changes being discussed among treebank developers concerns the inventory and design of the UD dependency relations.  \newcite{Croft2017LinguisticTM} proposed a redesign of the relations-inventory that would reflect four principles from linguistic typology while essentially not changing the topology of UD trees. By contrast, \newcite{gerdes-etal-2018-sud} proposed a surface-syntactic annotation scheme called SUD that would follow distributional criteria for defining the dependency tree structure and the naming of the syntactic functions. SUD results in changes to the dependency inventory and to tree topology.

Along another, more consensual line of exploration, proposals have been made regarding how to augment UD annotations so as to explicate additional predicate-argument relations between content words that are not captured by the basic surface-syntactic annotations of UD. \newcite{schuster-manning-2016-enhanced} formulated five kinds of enhancements, such as propagating the relation that the head of a coordination bears to the other conjuncts.
\newcite{candito-etal-2017-enhanced} added the neutralization of diathesis alternations as another kind of enhancement. Given that some of the enhancements require human annotations, \newcite{droganova-zeman-2019-towards} propose to identify a subset of deep annotations that can be derived semi-automatically from surface trees with acceptable quality.
The enhanced representation that results from these various proposals forms a directed graph but not necessarily a tree. It may contain ‘null’ nodes, multiple incoming edges and even cycles. Note that within the UD community, the enhancements are optional: it is acceptable for a treebank to be annotated with just a subset of possible enhancements. 

In order to highlight the interoperability of our proposed guidelines with other frameworks (such as SUD, \citep{gerdes-etal-2018-sud}), in Figure \ref{fig:nico-rosberg-UD} and in Figure \ref{fig:nico-rosberg-SUD} we display the same Italian tweet, represented in UD framework -- on the left -- and in SUD framework -- on the right. As it can be seen, in the text, a copula ellipsis occurs, as it is fairly common in news headlines and other kinds of user-generated content. The different approaches of UD and SUD, with regard to the election of syntactic heads and their dependents, or the different naming of syntactic functions, both pose no problems in the syntactic representation of such a case, therefore demonstrating the interoperability of our proposal.


\begin{figure}[!ht]
\begin{minipage}[b]{.45\textwidth}
\centering
\small
\begin{dependency}
\begin{deptext}
Nico  \&[.5cm] Rosberg  \&[.5cm] davanti  \&[.5cm] a  \&[.5cm] tutti \\
\end{deptext}
\depedge{1}{2}{flat:name}
\depedge[edge unit distance=2.5ex]{5}{1}{nsubj}
\depedge{3}{4}{fixed}
\depedge{5}{3}{case}
\deproot[edge unit distance=4ex]{5}{root}
\treetrans{`Nico Rosberg \textit{[was]} ahead of everyone'}
\end{dependency}
\caption{Italian example of copula ellipsis represented in UD. From POSTWITA, 2011.}\label{fig:nico-rosberg-UD}
\end{minipage}
\hfill
\begin{minipage}[b]{.45\textwidth}
\centering
\small
\begin{dependency}
\begin{deptext}
Nico  \&[.5cm] Rosberg  \&[.5cm] davanti  \&[.5cm] a  \&[.5cm] tutti \\
\end{deptext}
\depedge{1}{2}{flat@name}
\depedge[edge unit distance=4ex]{3}{1}{subj}
\depedge{3}{5}{comp:obj}
\depedge{3}{4}{unk@fixed}
\deproot[edge unit distance=4ex]{3}{root}
\treetrans{`Nico Rosberg \textit{[was]} ahead of everyone'}
\end{dependency}
\caption{Same example of ellipsis in an Italian tweet from Figure \ref{fig:nico-rosberg-UD}, but represented through the SUD framework.}\label{fig:nico-rosberg-SUD}
\end{minipage}
\end{figure}

\paragraph{Treatment of morphology: alignment with UniMorph}
Like the UD group, the collaborative UniMorph project \citep{Kirov16,kirov-etal-2018-unimorph,mccarthy-etal-2020-unimorph} is developing a cross-lingual schema for annotating the morphosyntactic details of language. While UD's main focus is on the annotation of dependency syntactic relations between tokens in corpora of running text, UniMorph focuses on the analysis of morphological features at the word type level. Nevertheless, UD corpora also include {more detailed}  annotations of lexical and grammatical properties of tokens beyond POS.

The attribute-value pairs used for morphological features in UD's schema are constructed in a bottom-up fashion: features are adapted and newly included as evidence comes in from languages for which resources are created. By contrast, UniMorph is top-down: its design is guided by surveys of the typological literature and aims to be complete, accounting for all attested morphological phenomena. Accordingly, UniMorph provides some attributes that UD lacks, such as those which describe information structure and deixis. In other cases, UniMorph provides more values for certain attributes: for instance, it covers 23 different noun classes used by Bantu languages.

As reported by \newcite{kirov-etal-2018-unimorph}, a preliminary survey of UD annotations shows that approximately 68\% of UD features  have direct UniMorph schema equivalents, with these feature sets covering 97.04\% of the complete UD tags.\footnote{\newcite{mccarthy-etal-2018-marrying} reports on  experiments in which UD feature annotation is deterministically converted to UniMorph format for multiple languages.} As the authors note, some UD features are outside the scope of UniMorph, which marks primarily morphosyntactic and morphosemantic distinctions. For example, UD has markers for abbreviated forms and foreign borrowings, which UniMorph does not provide as it limits its features to those needed for capturing the meanings of overt inflectional morphemes.  Conversely, some UniMorph features are not represented in UD due to its bottom-up approach.

While we present recommendations on user-generated content in this article, we neither extend the UPOS tagset nor the morphological features of the UD scheme. In that sense, existing mappings between UD and UniMorph are applicable to social media corpora. Nevertheless, our recommendations for the annotation of UGC go beyond the scope of UniMorph's top-down approach with new types of tokens, e.g. the morphological features of pictograms. 
As \newcite{mccarthy-etal-2020-unimorph} note, UniMorph now recognizes that lemmas and word forms can be segmented, hence in case of clitics or agglutinative formations, morphological features can be mapped onto segments of a word form. The resulting policy should be taken into account in aligning the current UD and UniMorph.

\section{Conclusion}
In this article we addressed the challenges of annotating user-generated texts from web and social media, 
proposing, in the context of Universal Dependencies, a unified scheme for their coherent treatment across different languages.

The variety and complexity of the UGC-related phenomena sometimes renders as non-trivial their adequate representation by means of an already existing scheme, such as UD.
We hope that this proposal will trigger discussions throughout the treebanking community and will pave the way for a uniform handling of user-generated content in a dependency framework.

\section*{Acknowledgements}
 We warmly thank Nathan Schneider for his comments on a previous version of this paper. All remaining errors are ours. The work of C. Bosco is partially funded by Progetto di Ateneo/CSP 2016 ({\em Immigrants, Hate and Prejudice in Social Media}, S1618\_L2\_BOSC\_01). The work of M. Sanguinetti is funded by PRIN 2017 (2019-2022) project \textit{HOPE - High quality Open data Publishing and Enrichment}.  Ö. Çetinoğlu is funded by DFG via project CE 326/11 \textit{Computational Structural Analysis of German Turkish Code-Switching (SAGT)}. D. Seddah is partially funded by  the ANR projects ParSiTi (ANR-16-CE33-0021) and SoSweet (ANR15-CE38-0011-01). T. Lynn and L. Cassidy are funded by the Irish Government Department of Culture, Heritage and the Gaeltacht under the GaelTech Project, and also supported by Science Foundation Ireland in the ADAPT Centre (Grant 13/RC/2106) at Dublin City University.


%
%


\newpage

\pagestyle{empty}

\newgeometry{left=40mm,
             right=30mm,
             top=6mm,
             bottom=6mm}

\begin{landscape}
\section*{Appendix}

\begin{table}[h!]
\centering
\tiny
\begin{tabular}{l|l|cc|cc|cc|c|cc|ccc}
\multicolumn{2}{c}{\textbf{}}  & \multicolumn{2}{c}{\textbf{TOKEN}} & \multicolumn{2}{c}{\textbf{LEMMA}} & \multicolumn{2}{c}{\textbf{UPOS}} & {\textbf{FEATS}} & \multicolumn{2}{c}{\textbf{DEPREL}} & {\textbf{MISC}} \\
 \toprule
{\textbf{ANNOTATION ISSUE}} &  & \textbf{No change} & \textbf{Split} & \textbf{No change} & \textbf{Normalize} & \textbf{Standard synt. role} & \textbf{Other} &  & \textbf{Standard synt. role} & \textbf{Other} &  &  \\

\midrule

{Diacritic Omission} & & \checkmark &  &  & \checkmark & \checkmark &  &  & \checkmark &  &  \texttt{NonCan=SpellVar|CorrectForm} \\

{Vowel Omission} & & \checkmark &  &  & \checkmark & \checkmark &  &  & \checkmark &  &  \texttt{NonCan=CharOm|CorrectForm} \\

{Phonetization} & & \checkmark &  &  & \checkmark & \checkmark &  &  & \checkmark &  &  \texttt{NonCan=Phon|CorrectForm} \\

{Spelling Errors} & & \checkmark &  &  & \checkmark & \checkmark &  & \texttt{Typo=Yes} & \checkmark &  &  \texttt{CorrectForm} \\

{Abbreviation} & & \checkmark &  &  & \checkmark & \checkmark &  & \texttt{Abbr=Yes} & \checkmark &  &  \texttt{FullForm} \\

\midrule

\multirow{3}{*}{Contraction} & Canonical &  & \checkmark &  & \checkmark & \checkmark &  & \texttt{Abbr=Yes} & \checkmark &  & \texttt{NonCan=Cont} \\
 & Noncanonical\&Unintentional &  & \checkmark &  & \checkmark & \checkmark &  & \texttt{Typo=Yes} & \checkmark &  & \texttt{CorrectForm|CorrectSpaceAfter|SpaceAfter|NonCan=Cont}\\
 & Noncanonical\&Intentional & \checkmark &  & \checkmark &  & \checkmark &  & \texttt{Abbr=Yes} & \checkmark &  & \texttt{NonCan=Cont} \\

\midrule
 
{Oversplitting} & & \checkmark &  & \checkmark {(remaining tokens)} & \checkmark {(first token)} & \checkmark {(first token)} & {\texttt{X} (remaining tokens)} & & \checkmark {(first token)} & \texttt{goeswith} & \texttt{NonCan=OS}  \\

{Punctuation Reduplication} & & \checkmark &  &  & \checkmark & \checkmark &  &  & \checkmark &  &   \texttt{NonCan=PunctVar} &  \\
{Graphemic Stretching} & & \checkmark &  &  & \checkmark & \checkmark &  &  & \checkmark &  &  \texttt{CorrectForm|NonCan=Stretch} \\

{Disguise} & & \checkmark &  &  & \checkmark & \checkmark &  &  & \checkmark &  & \texttt{CorrectForm|NonCan=SpellVar} \\

{Transliteration} & & \checkmark &  &  & \checkmark (remove inflection) & \checkmark &  &  & \checkmark &  & \texttt{CorrectForm|NonCan=Transl} \\

{Lexical Innovation} & & \checkmark &  &  &  & \checkmark & \checkmark (remove inflection) &  & \checkmark &  & \texttt{NonCan=LexInno}  \\

{Truncation} & & \checkmark &  & \checkmark &  & \checkmark (if known) & \texttt{X}(if not known) &  & \checkmark &  & \texttt{FullForm|NonCan=Trunc} (if known) \\

{Autocorrection} &  & \checkmark &  &  & \checkmark & \checkmark (if known) & \texttt{X}(if not known) &  & \checkmark &  & \texttt{CorrectForm|NonCan=AutoC} \\

\midrule

\multirow{2}{*}{Hashtags} & synt. integrated & \multirow{2}{*}{\checkmark} &  & \multirow{2}{*}{\checkmark} &  &  & \multirow{2}{*}{\texttt{X}} &  & \checkmark &  & \texttt{FuncPOS} \\
 & standalone &  &  &  &  &  &  &  &  & \texttt{parataxis:hashtag} &  \\

\midrule

\multirow{2}{*}{At-mentions} & synt. integrated & \multirow{2}{*}{\checkmark} &  & \multirow{2}{*}{\checkmark} &  &  & \multirow{2}{*}{\texttt{PROPN}} &  & \checkmark &  &  &  &  \\
 & standalone &  &  &  &  &  &  &  &  & \texttt{vocative:mention} &  \\
 
\midrule

\multirow{2}{*}{URLs} & synt. integrated & \multirow{2}{*}{\checkmark} & \multirow{2}{*}{} & \multirow{2}{*}{\checkmark} &  &  & \multirow{2}{*}{\texttt{SYM}} &  & \checkmark &  &  \\
 & standalone &  &  &  &  &  &  &  &  & \texttt{parataxis:url} &  \\

\midrule

\multirow{2}{*}{Pictograms/Emoticons} & synt. integrated & {\multirow{2}{*}{\checkmark (single)}} & {\multirow{2}{*}{\checkmark (multiple)}} & \multirow{2}{*}{\checkmark} & &  & \multirow{2}{*}{\texttt{SYM}} &  & \checkmark   &  &  \\
 & standalone &  &  &  &  &  &  &  &  & \texttt{discourse} &  \\

\midrule

\multirow{2}{*}{RTs} & synt. integrated & \multirow{2}{*}{\checkmark} &  &  & \checkmark (remove inflection) & \checkmark (\texttt{VERB}/\texttt{NOUN}) &  & \texttt{Abbr=Yes} & \checkmark &  &  \\
 & standalone &  &  & \checkmark &  &  & \texttt{SYM} &  &  & \texttt{parataxis} & \texttt{FullForm} \\

\midrule

Markup symbols &  & \checkmark &  & \checkmark &  &  & \texttt{SYM} &  &  & \texttt{punct} &  \\

\midrule

\multirow{2}{*}{Code-switching} & INTRA & {\multirow{2}{*}{\checkmark} } &  & {\multirow{2}{*}{\checkmark (if not known)}} & \multirow{2}{*}{\checkmark (if known)} & {\multirow{2}{*}{\checkmark (if known)}} & {\multirow{2}{*}{\texttt{X} (if not known)}} & \multirow{2}{*}{\texttt{Foreign=Yes}} & \checkmark &  & \texttt{CSType=INTRA|LangID={[}isocode{]}} \\
 & INTER &  &  &  &  &  &  &  & &\texttt{flat}/\texttt{flat:foreign} & \texttt{CSType=INTER|LangID={[}isocode{]}} \\

\midrule

\multirow{2}{*}{Disfluencies} & Repairs & {\multirow{2}{*}{\checkmark}} &  & {\multirow{2}{*}{\checkmark}} &  & {\multirow{2}{*}{\checkmark (if known)}} & {\multirow{2}{*}{\texttt{X} (if not known)}} &  &  & \texttt{reparandum}  &  \\
 & Hesitation markers &  &  &  &  &  & &  &  & \texttt{discourse}  &  \\

\midrule
 
{Sentence boundaries} &  &  &  &  &  &  &  &  & \texttt{parataxis:sentence} &  \\
\bottomrule
\end{tabular}\caption{{Summary of CoNLL-U proposed implementations.}}
\end{table}
\end{landscape}

\restoregeometry

\bibliographystyle{plainnat}
\bibliography{bibfromlrec}

\end{document}